\documentclass[journal]{IEEEtran}
\usepackage{amsmath, amsfonts}
\usepackage{algorithmic}
\usepackage{algorithm}
\usepackage{array}
\usepackage[caption=false, font=normalsize, labelfont=sf, textfont=sf]{subfig}
\usepackage{textcomp}
\usepackage{stfloats}
\usepackage{url}
\usepackage{verbatim}
\usepackage{graphicx}
\usepackage{cite}
\usepackage[justification=centering]{caption}
\usepackage{ragged2e}
\usepackage{tabularx}
\usepackage{graphicx}  
\usepackage{float} 
\usepackage{makecell}
\usepackage{longtable}
\usepackage{booktabs}
\usepackage{textcase}
\usepackage{multirow}
\usepackage{multicol}
\usepackage{xcolor} 
\usepackage{tikz, xcolor}

\usepackage[implicit=true]{hyperref}
\usepackage{colortbl}

\hypersetup{hidelinks, 
	colorlinks=true, 
	allcolors=blue, 
	pdfstartview=Fit, 
	breaklinks=true}

\usepackage[numbers, sort&compress]{natbib}

\definecolor{lime}{HTML}{A6CE39}
\DeclareRobustCommand{\orcidicon}{
\begin{tikzpicture}
\draw[lime, fill=lime] (0, 0)
circle[radius=0.16]
node[white]{{\fontfamily{qag}\selectfont \tiny \.{I}D}};
\end{tikzpicture}
\hspace{-2mm}
}
\foreach \x in {A, ..., Z}{%
\expandafter\xdef\csname orcid\x\endcsname{\noexpand\href{https://orcid.org/\csname orcidauthor\x\endcsname}{\noexpand\orcidicon}}
}

\newcommand{\romannumber}[1]{\romannumeral#1}
\newcommand{\majorrevision}{\color{black}}
\newcommand{\minorrevision}{\color{black}}
\newcommand{\thirdrevision}{\color{black}}
\newcommand{\thirdrevisiont}{\cellcolor{gray!20}\color{black}}

\begin{document}

\title{Detect Changes like Humans: Incorporating Semantic Priors for Improved Change Detection}

\author{
Yuhang Gan,
Wenjie Xuan,
Zhiming Luo,
Lei Fang,
Zengmao Wang,~\IEEEmembership{Member,~IEEE,} \\
Juhua Liu,~\IEEEmembership{Member,~IEEE,}
Bo Du,~\IEEEmembership{Senior Member,~IEEE}

\thanks{This work was supported in part by the National Natural Science Foundation of China under Grants U23B2048 and 62225113, in part by the Science and Technology Major Project of Hubei Province under Grants 2024BAB046 and 2025BCB026, and in part by Major Special Project of China Innovation Challenge (Ningbo) under Grant 2024T008. The numerical calculations in this paper have been done on the supercomputing system in the Supercomputing Center of Wuhan University. \textit{Yuhang Gang and Wenjie Xuan contributed equally to this work. Corresponding authors: Zengmao Wang, Juhua Liu (e-mail: \{wangzengmao, liujuhua\}@whu.edu.cn).}}

\thanks{Yuhang Gan, Wenjie Xuan, Zhiming Luo, Zengmao Wang, Juhua Liu and Bo Du are with the School of Computer Science, Wuhan University, Wuhan, China. Y. Gan is also with Land Satellite Remote Sensing Application Center, MNR, Beijing, China (e-mail: \{ganyuhang, dreamxwj, thislzm, wangzengmao, liujuhua, dubo\}@whu.edu.cn).}

\thanks{Lei Fang is with CAAZ(Zhejiang) Information Technology Co., Ltd., Ningbo, China. (e-mail: icedark@zju.edu.cn).}
}




\maketitle

\begin{abstract}
    When given two similar images, humans identify their differences by comparing the appearance ({\it e.g., color, texture}) with the help of semantics ({\it e.g., objects, relations}). However, mainstream {\majorrevision binary} change detection models adopt a supervised training paradigm, where the annotated binary change map is the main constraint. Thus, such methods primarily emphasize difference-aware features between bi-temporal images, and the semantic understanding of changed landscapes is undermined, resulting in limited accuracy in the face of noise and illumination variations. 
    To this end, this paper explores incorporating semantic priors from visual foundation models to improve the ability to detect changes. Firstly, we propose a Semantic-Aware Change Detection network~(SA-CDNet), which transfers the knowledge of visual foundation models ({\it i.e., FastSAM}) to change detection. Inspired by the human visual paradigm, a novel dual-stream feature decoder is derived to distinguish changes by combining semantic-aware features and difference-aware features. {\thirdrevision Secondly, we explore a single-temporal pre-training strategy for better adaptation of visual foundation models. With pseudo-change data constructed from single-temporal segmentation datasets, we employ an extra branch of proxy semantic segmentation task for pre-training. We explore various settings like dataset combinations and landscape types, thus providing valuable insights.}
    Experimental results on five challenging benchmarks demonstrate the superiority of our method over the existing state-of-the-art methods. 
    The code is available at \href{https://github.com/DREAMXFAR/SA-CDNet}{SA-CD}.
\end{abstract}


\section{Introduction}
\label{Introduction}
    
    \begin{figure}[ht]
        \centering
        \includegraphics[width=0.98\linewidth]{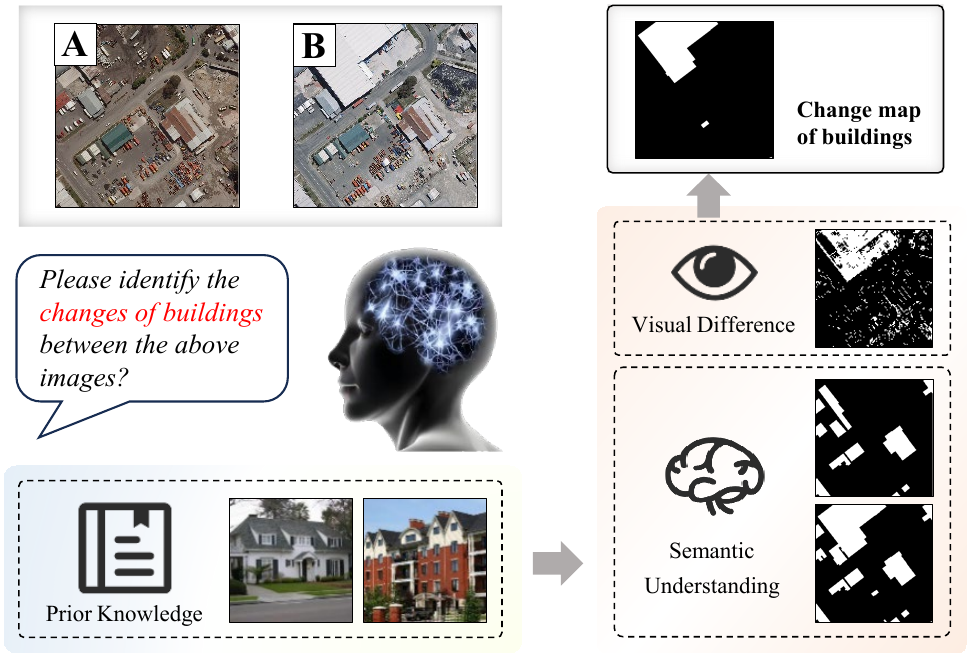}
        \caption{\justifying To locate changes between images, humans not only perceive appearance differences but also utilize semantic priors unconsciously for help. For instance, humans would identify subjects in each image via prior knowledge and then compare them, which reflects the importance of semantics in detecting changes. However, current {\majorrevision binary} change detection methods are inclined to learn difference-aware features through the target difference map for training, but they ignore semantic understandings and bring sub-optimal results.}
        \label{fig:motivation}
        \vspace{-0.5cm}
    \end{figure}
    
    \IEEEPARstart{R}{emote} sensing change detection identifies interested landscape changes and interprets variations from bi-temporal images covering the same region~\cite{wu2024unet, 10216355, 10833830}. It plays a crucial role in environmental monitoring, natural disaster assessment, urban planning, {\it etc}. To handle change detection with massive high-resolution~\cite{cnn1, cnn2, changeformer, transformer1} {\majorrevision{and hyperspectral~\cite{10216355, 10833830}}} remote sensing images effectively and efficiently, deep-learning-based methods have grown prosperity for their powerful representation capability and impressive performance in recent years. {\majorrevision This paper focuses on the detection of binary changes for high-resolution remote sensing images.}

    Existing deep-learning-based {\majorrevision binary} change detection methods~\cite{ chen2022semantic, 10049565, wang2023changes, LIU2022108960} mostly employed a siamese architecture with supervision, consisting of a network to extract features and a classifier to discriminate changed regions. The mainstream methods adopted CNNs~\cite{cnn1,cnn2} or Transformers~\cite{changeformer,transformer1,mctnet2023} as the feature extractor to provide discriminative representations for bi-temporal images. The classifier usually selects a fully connected or softmax layer to backpropagate loss. The main supervision is a binary difference map that represents the interested subject changes between bi-temporal images. Previous methods~\cite{chen2022semantic, 10049565, wang2023changes} aim to learn an informative feature space where the features of changed pixels are pushed apart and those unchanged ones are pulled together. Since the binary difference map only provides the change information, such methods are inclined to learn difference-aware change features. Therefore, {\it they would fall short of learning semantic-aware features, which deviates from the human visual paradigm and leads to sub-optimal results}. As illustrated in Fig.~\ref{fig:motivation}, when given two images of the same area, humans find the changed regions encouraged by both the appearance difference and semantics~\cite{tu2021semantic}. Generally, the appearance differences represent the absolute visual changes between the two images. Such differences would be caused by multiple factors (\textit{e.g.}, seasonal or illumination variations) apart from the true landscape changes. The semantics are utilized unconsciously, but it is helpful and vital to distinguish the interested subject changes like buildings and farmland. Motivated by this, we explore incorporating richer semantic priors for improved change detection. 

    This paper proposes to enhance the semantic understanding of changed landscapes from three aspects: {\bf \romannumber{1})} We exploit visual foundation models ({\it i.e., FastSAM}~\cite{fastSAM}) to encode features and adapt them to change detection via an adapter. This aims to endow the network with knowledge of foundation models, thus providing rich semantic priors. {\bf \romannumber{2})} Inspired by the human visual paradigm, a dual-stream feature decoder is proposed to extract difference-aware features and semantic-aware features individually. Such features are proven to be complementary in our experiments, especially when pixel-level variations exist. Thus, we propose an advanced network for change detection, namely SA-CDNet. {\thirdrevision {\bf \romannumber{3})} To further enrich the semantic understanding of landscapes, we adopt a single-temporal pre-training strategy following \cite{zheng2021change} for simplicity. Specifically, we construct large-scale pseudo-change detection data with remote sensing segmentation data, and introduce an extra branch for semantic segmentation as an additional pre-training task. We demonstrate the effectiveness and scalability of \cite{zheng2021change} by conducting detailed discussions on landscape types, different datasets, and combinations for pre-training, offering insights for adapting foundation models to change detection.}

    Our main contributions can be summarized as follows:
    \begin{itemize}
        \item We propose an advanced SA-CDNet for {\majorrevision binary} change detection, which inherits the rich knowledge of natural images from the visual foundation model and adapts semantic priors to change detection. Besides, a dual-stream feature decoder is derived to mimic the human visual paradigm, comprising a semantic-aware decoder and a difference-aware decoder for robust predictions. 

        \item {\thirdrevision We adopt a single-temporal semantic pre-training strategy with an extra proxy segmentation branch and prove its effectiveness on better adapting foundation models to change detection tasks. The extensive ablations on various pre-training settings, including datasets and landscape types, provide valuable insights for the community.}

        \item Extensive experiments are conducted on five challenging benchmarks, \textit{i.e.},  LEVIR-CD, LEVIR-CD+, S2Looking, WHU-CD for building changes, and WHU Cultivated Land for farmland changes. Our method achieves SOTA performance on all datasets. We also ablate the modules and discuss different settings for the pre-training in detail to prove their effectiveness. 
    \end{itemize}

    In the rest of paper, Sec.~\ref{Related work} reviews related works. Sec.~\ref{Our Method} introduces our SA-CDNet and pre-training strategy. Sec.~\ref{Experimental results} includes extensive experiments and ablation studies on five benchmarks. Finally, Sec.~\ref{Conclusion} concludes the paper.  

\section{Related Work}
\label{Related work}
    \subsection{Remote Sensing Change Detection}
    \label{Change Detection in Remote Sensing}
        Change detection is a long-standing topic in remote sensing. 
        Recent improvements of change detection techniques were mainly attributed to the advanced deep-learning methods~\cite{wu2024unet}. The pioneer work~\cite{daudt2018fully} proposed a siamese structure based on U-Net, which leveraged CNNs to capture local spatial context and yielded attractive performance. To cover the information loss during the successive down-sampling in CNNs, SNUNet~\cite{snunet} introduced dense skip connections and multi-scale feature fusion to enhance representations for detecting detailed changes. ViT~\cite{dosovitskiy2020image} was introduced to change detection due to its global context representations. SwinSUNet~\cite{zhang2022swinsunet} constructed a fully transformer-based encoder-decoder network based on Swin Transformer. Similarly, BIT~\cite{bit} introduces the self-attention mechanism to capture global features, thereby enhancing change detection performance.

        The above methods primarily focus on enhancing representations through advanced network architectures or attention mechanisms to realize increments. However, given that only binary difference maps are available for supervised training in change detection tasks, existing studies focused on difference-aware features and overlooked the enhancement of semantic understanding. We argue that this deviates from the human visual paradigm, where both pixel difference and semantics are crucial in distinguishing subject changes. Therefore, this paper primarily explores the incorporation of richer semantic priors into change detection for improvements.

        {\majorrevision Although semantic change detection approaches~\cite{yang2020semantic,peng2021scdnet,zhou2022multi,ding2022bi} also incorporate landscape semantics to detect changes, sharing some similarity to our motivation, we claim that it is a different setting. The semantic change detection assumes that the semantic maps of both bi-temporal images are always accessible, which provides additional supervision and regularization for training. However, only the difference map is available in binary change detection, making it more challenging to exploit semantic information from bi-temporal images during training. To this end, we introduce visual foundation models and exploit a single-temporal semantic pre-training strategy with pseudo-change data to enrich landscape semantics.}
    
    \subsection{Visual Foundation Models}
    \label{foundation model}
        Recently, visual foundation models~\cite{bommasani2021opportunities} have raised great attention in the computer vision community. Through training models with billions of parameters on extensive data, these models not only demonstrate promising performance but also exhibit unparalleled capability of image understanding and zero-shot generalization to unseen subjects. The representative SAM~\cite{sam} can segment any object of interest with high accuracy via prompts. Inspired by its great success, many variants \cite{xiong2024efficientsam, mobileSAM, fastSAM} with lightweight encoders have been proposed, making it more efficient and easier to deploy. {\thirdrevision Recent studies endeavored to integrate SAM for change detection, leveraging its powerful representations and semantic understandings for advanced performance. Early attempts directly introduce SAM features~\cite{liu2024laddering,zhao2024fine,wei2025ass} or segmentation predictions~\cite{sun2024segment,qiu2024ded,gao2025combining} to existing change detectors, thus providing semantic priors for guidance. Subsequent works ~\cite{altam2023samcd,samcd,chen2024time,zhang2024integrating,wei2025siamese,kong2024fastsam} focus on adapting SAM to change detection. Commonly, they would construct a siamese architecture with the pre-trained SAM encoder to extract bi-temporal image features, which are then decoded for change prediction. To bridge the domain gap between SAM's training objective and change detection, researchers employ LoRA variants~\cite{chen2024time,zhang2024integrating} or lightweight adapters~\cite{altam2023samcd,wei2025siamese,li2025sam} to finetune the SAM encoder with change detection data. Recent approaches~\cite{samcd,kong2024fastsam,gao2025combining} favors efficient alternatives, {\it e.g.,} FastSAM~\cite{fastSAM} and MobileSAM~\cite{mobileSAM} to relieve SAM's high computational cost. Despite a little sacrifice in performance, these methods are more efficient and application-friendly.}
        
        {\thirdrevision Our work distinguishes from previous works in two aspects: 1) While existing approaches focus on parameter-efficient finetuning techniques for adaptation, the design of feature decoders is under-explored, where common practice includes single CNNs~\cite{altam2023samcd,samcd,chen2024time,zhang2024integrating,wei2025siamese} or UNet decoders~\cite{wei2025ass,gao2025combining,kong2024fastsam}. We propose a more powerful dual-stream decoder to better inherit FastSAM's semantic understanding for change detection. 2) We also uncover the effectiveness of domain-specific pre-training for change adapters and decoders by exploring various pre-training settings. Our work is an extension of the influential SAM-CD~\cite{samcd}, and it features effective dual-stream feature decoders and the introduction of single-temporal pre-training. Experiments in Sec.~\ref{sec:ablation} prove the advance of our designs.}

\section{Methodology}
\label{Our Method}
    This section first reveals the detailed structure of our SA-CDNet in Sec.~\ref{SANet}. Sec.~\ref{stsps} introduces the single-temporal semantic pre-training strategy, including the pseudo-change detection data construction and the proxy segmentation task design. The overall ``pre-training + fine-tuning" pipeline is concluded in Sec.~\ref{strategy}. 

    \subsection{Semantic-aware Change Detection Network} 
    \label{SANet}
    
\begin{figure*}[ht!]
            \setlength{\abovecaptionskip}{-10pt} \setlength{\belowcaptionskip}{-1pt}
            \begin{center}
            \includegraphics[width=0.98\textwidth]{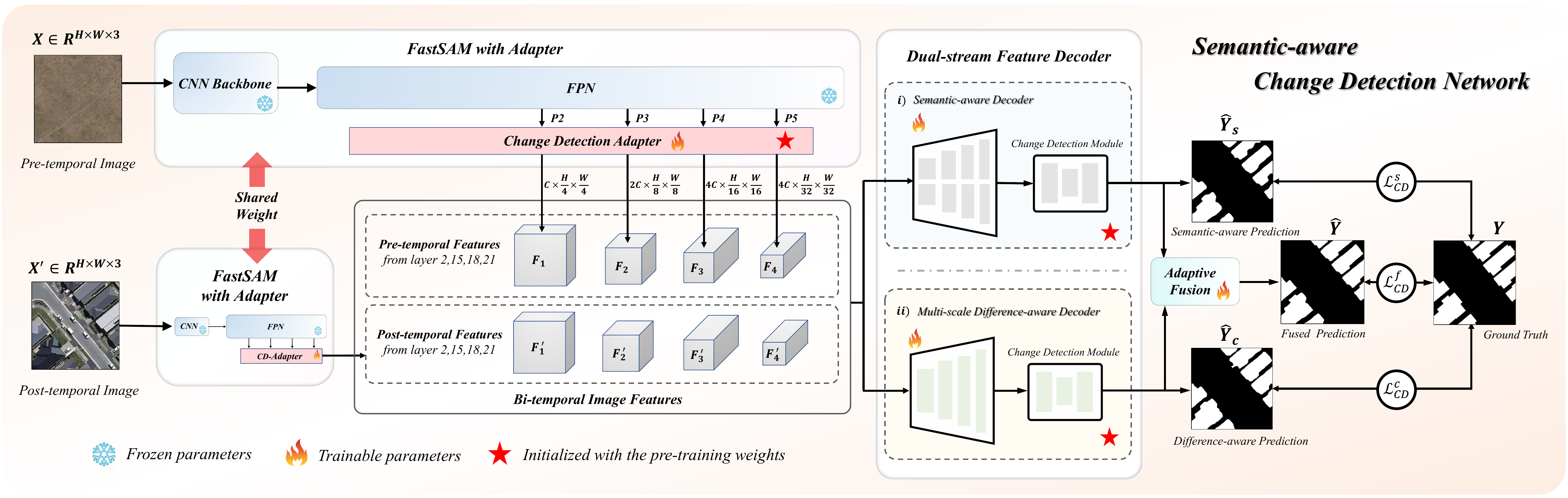}
            \end{center}
            \vspace{0.4cm}
            \caption{\justifying The overview of SA-CDNet. We employ the frozen FastSAM encoder to extract features, then refined by an adapter to align features for change detection task. The dual-stream feature decoder contains a semantic-aware feature decoder and a multi-scale difference-aware feature decoder, providing richer semantic-aware and difference-aware features. The adaptive fusion module combines such features and produces the final change map. 
            Note that the adapter and dual-stream decoder are initialized with the single-temporal semantic pre-training introduced in Sec.~\ref{stsps}.}            
            \label{fig:sa-cdnet}
            \vspace{-0.4cm}  
        \end{figure*}

        The overall architecture of SA-CDNet is presented in Fig.~\ref{fig:sa-cdnet}, primarily consisting of three components: {\bf \romannumber{1})} A visual foundation model-based feature encoder ({\it i.e., FastSAM}) with a change detection adapter to provide discriminative representations. {\bf \romannumber{2})} A dual-stream feature decoder, containing a semantic-aware decoder branch and a multi-scale difference-aware branch, to decode semantic-aware and difference-aware features for human-like change detection. {\bf \romannumber{3})} An adaptive fusion module to combine the semantic-aware and difference-aware features for change predictions. 

        When given bi-temporal images $X, X^{\prime} \in \mathbb{R}^{H \times W\times 3}$, the network aims to produce a binary change map $\hat{Y} = \{y_{ij}  \in\{0,1\}\}^{H \times W}$ indicating changed regions. ${Y}$ denotes the ground-truth change map. Specifically, the visual foundation model-based feature encoder $\mathcal{E}(\cdot)$ first extracts features from the two images separately at the FPN~\cite{Lin_2017_CVPR}, denoted as $\{P_i, P^{\prime}_i\}_{i=2, 3, 4, 5}$. Since the vision foundation model is commonly pre-trained on extensive natural scene image datasets, it keeps rich semantic priors of natural images. We freeze its parameters to keep its inner knowledge unaffected when adapting it to change detection. Besides, similar to SAM-CD~\cite{samcd}, an adapter $\mathcal{A}(\cdot)$ is introduced to align features, i.e., $\{F_i, F^{\prime}_i\}_{i=2, 3, 4, 5}$, to remote sensing change detection. These refined features would be fed to the dual-stream decoder to further mine semantic-aware features $\{S_i, S^{\prime}_i\}_{i=2, 3, 4, 5}$ through the semantic-aware decoder branch $\mathcal{M}(\cdot)$ and difference-aware features $\{D_i, D^{\prime}_i\}_{i=2, 3, 4, 5}$ through the multi-scale difference-aware decoder branch $\mathcal{N}(\cdot)$ independently. Finally, the adaptive fusion module would complement difference-aware features with the semantic-aware features and predict the change map $\hat{Y}$ with a softmax classifier. The ground truth $Y$ is used to compute the loss. For a better understanding, we elaborate on the detailed structure and designs of each module as follows. 

        \subsubsection{\textbf{Visual foundation model-based feature encoder with change detection adapter}}
        \label{model: encoder}
            Recently, visual foundation models like SAM~\cite{sam} and its variants~\cite{seem2023,xiong2024efficientsam,mobileSAM,fastSAM,zhu2024medical} have shown impressive performance and zero-shot ability on semantic segmentation and displayed profound semantic understandings of subjects in the natural images. To marry the rich semantic priors from visual foundation models to change detection, we borrow the encoder of visual foundation models to obtain discriminative features from bi-temporal images first. We compared SAM and its three variants, including EfficientSAM~\cite{xiong2024efficientsam}, FastSAM~\cite{fastSAM}, and MobileSAM~\cite{mobileSAM} in Sec.~\ref{sec:ablation}, and finally selected FastSAM-x as the encoder for a trade-off between performance and efficiency. 
            We highlight that {\it the FastSAM encoder can be substituted by that of another foundation model} because we only employ a frozen encoder to extract bi-temporal image features. Specifically, the FastSAM feature encoder inherits Yolo-v8 structure, which comprises a CNN backbone with FPN to extract features. We input bi-temporal images separately and select the features from the $2^{nd}$, $15^{th}$, $18^{th}$, and $21^{st}$ layers ({\it i.e.}, the final layer of each stage) as $\{P_i, P^{\prime}_i\}_{i=2, 3, 4, 5}$. We freeze the parameters of FastSAM to avoid negative impacts on its inner knowledge. 
            
            Considering the semantic gap between natural images and remote sensing images, we follow SAM-CD~\cite{samcd} and introduce an adapter $\mathcal{A}(\cdot)$ for feature alignment as shown in Fig.~\ref{fig:sa-cdnet}, thus adapting representations to change detection. The adapter contains a $1\times 1$ convolution, batch normalization, and ReLU function. The process of feature alignment can be denoted as:
            \begin{equation}
                F_i = A(P_i) = ReLU\{BN[Conv(P_i)]\}, 
            \end{equation}
            where $P_i$ is the selected features of FastSAM and $F_i$ is the aligned feature. The adapter fits the features from the frozen foundation model to change detection.
            
        \subsubsection{\textbf{Dual-stream feature decoder}}
        \label{model: decoder}        
            Motivated by the fact that humans rely on both the appearance difference and semantics to identify changed regions, we design a dual-stream feature decoder to mine semantic-aware features and difference-aware features separately for improved performance. The detailed structures of the two branches are revealed as follows. 

\begin{figure}[]
                \setlength{\abovecaptionskip}{-10pt} \setlength{\belowcaptionskip}{-1pt}
                \begin{center}
                \includegraphics[width=0.47\textwidth]{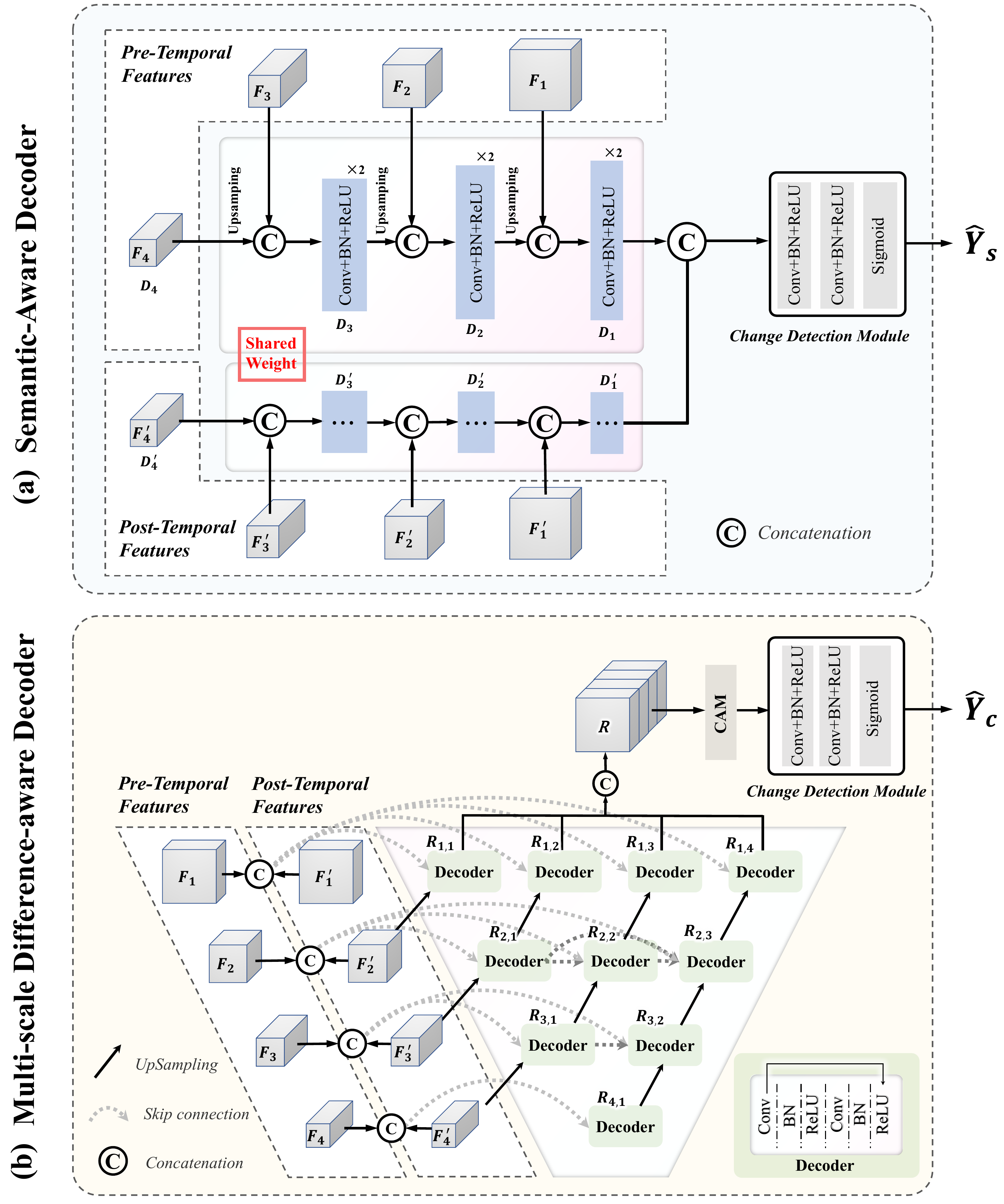}    
                \end{center}
                \vspace{0.4cm}
                \caption{\justifying The structure of dual-stream feature decoder, which includes (a) the semantic-aware decoder and (b) the multi-scale difference-aware decoder. }
                \label{SAD}
                \vspace{-0.2cm}
            \end{figure}

            {\it Semantic-aware decoder} $\mathcal{M}(\cdot)$ employs a simple network that contains three decoding units to decode bi-temporal features individually. As shown in Fig.~\ref{SAD}(a), each decoding unit comprises an up-sampling and two \texttt{conv + batch norm + ReLU} blocks, progressively fusing multi-scale features of single-temporal images. Taking unit $i$ for example, it receives decoded features $D_{i+1}$ from the previous stage, performs deconvolution to upsample $D_{i+1}$, then concatenates it with the current feature $F_{i}$ and feeds to the decoder unit. The extracted features $\{F_i, F^{\prime}_i\}_{i=2, 3, 4, 5}$ come from four scales $(\frac{H}{4k}, \frac{W}{4k})_{k=1, 2, 4, 8}$, which is expressed as{\majorrevision :} 
            \begin{equation}
                \label{2-1}
                D_{i} = Conv[F_{i}, \text{Upsample}(D_{i+1})],\ \ D_4 = F_4.
            \end{equation} 
            Then the decoded features $D_4, D^\prime_4$ are further concatenated and fed to a change detection module that consists of two convolutional blocks and a sigmoid layer for change prediction. Thus, the overall computation of the semantic-aware decoder branch is formulated as{\majorrevision :}
            \begin{equation}
               \hat{Y}_s = \mathcal{M}(\{F_i, F^\prime_i\}_{i=2, 3, 4, 5}),
            \end{equation}
            where $\hat{Y}_s$ is the output semantic-aware change map. Since bi-temporal image features are extracted and fused independently, we suppose this design helps to keep the semantics of single-temporal images. Besides, the late fusion of bi-temporal image features implicitly exploits the landscape semantics to distinguish changed regions. As validated in Sec.~\ref{sec:ablation}, the semantic-aware decoder is advanced to capture subject semantics and is robust to seasonal and illumination variations. 

            {\it Multi-scale difference-aware decoder} $\mathcal{N}(\cdot)$ takes a densely connected structure as presented in Fig.~\ref{SAD}(b). The basic decoding unit consists of two \texttt{conv + batch norm + ReLU} blocks. Given the extracted $\{F_i, F^{\prime}_i\}_{i=2, 3, 4, 5}$, the feature maps of the same scale are first concatenated and then delivered to the densely-connected decoding units, in order to mine difference-aware features. {\minorrevision Following a UNet++ style~\cite{snunet,peng2019end}}, each unit also receives features from adjacent scales to enrich multi-scale representations. This process can be expressed as{\majorrevision :} 
            \begin{equation}
                \label{3-5}
                R_{i, j} = \begin{cases}
                Conv[F_i, F_i^{\prime}, \text{Up}(F_i^{\prime})] & j = 1,  \\
                Conv[F_i, F_i^{\prime}, [R_{i, k}]_{k}^{j-1}, \text{Up}(R_{i+1, j-1})] & j > 1, 
                \end{cases}
            \end{equation}
            where $R_{i, j}$ denotes the output of the $(i, j)$ decoding unit. $F_i, F_i^{\prime}$ are pre- and post-temporal image features. $\text{Up}$ means the upsampling with deconvolution. After the decoding process, the fused multi-scale features $\{R_{1, j}\}_{j=1,2,3,4}$ are collected. We employ a Channel Attention Module~(CAM)~\cite{cam} to integrate all difference information, and use a change detection module with the sigmoid function to predict the difference-aware change map $\hat{Y}_c$. The overall computation of multi-scale difference-aware decoder branch is formulated as{\majorrevision :}  
            \begin{equation}
                \label{3-6}
                \hat{Y}_c = \mathcal{N}(\{F_i, F^\prime_i\}_{i=2, 3, 4, 5}). 
            \end{equation}
            In contrast to the semantic-aware decoder that deals with bi-temporal image features individually, this difference-aware decoder mines difference-aware features from rich multi-scale features of bi-temporal images, thus helping to discriminate change regions as validated in Sec.~\ref{sec:ablation}. It is sensitive to changes of various scales and handles details effectively. 

        \subsubsection{\textbf{Adaptive fusion module}}
        \label{model: fusion} 
            As mentioned, the semantic-aware decoder excels at capturing semantic-aware and contextual features, and the multi-scale difference-aware decoder prevails in perceiving fine-grained difference-aware features. We employ a simple adaptive fusion module to incorporate the difference-aware features and semantic-aware features for final results. For simplicity, we introduce a learnable weight $\sigma(\omega)$ to integrate the predictions from the two decoders, where $\sigma(\cdot)$ denotes the sigmoid function. This fusion strategy obtained a better trade-off between the precision and recall as discussed in Tab.~\ref{tab:fusion}. Thus, the final prediction $\hat{Y}$ is computed as{\majorrevision :} 
            \begin{equation}
                \label{adf}
                \hat{Y} = \sigma(\omega) * \hat{Y}_s + (1 - \sigma(\omega)) * \hat{Y}_c. 
            \end{equation}

        \begin{figure*}[ht!]
            \setlength{\abovecaptionskip}{-10pt} \setlength{\belowcaptionskip}{-1pt}
            \begin{center}
            \includegraphics[width=0.98\textwidth]{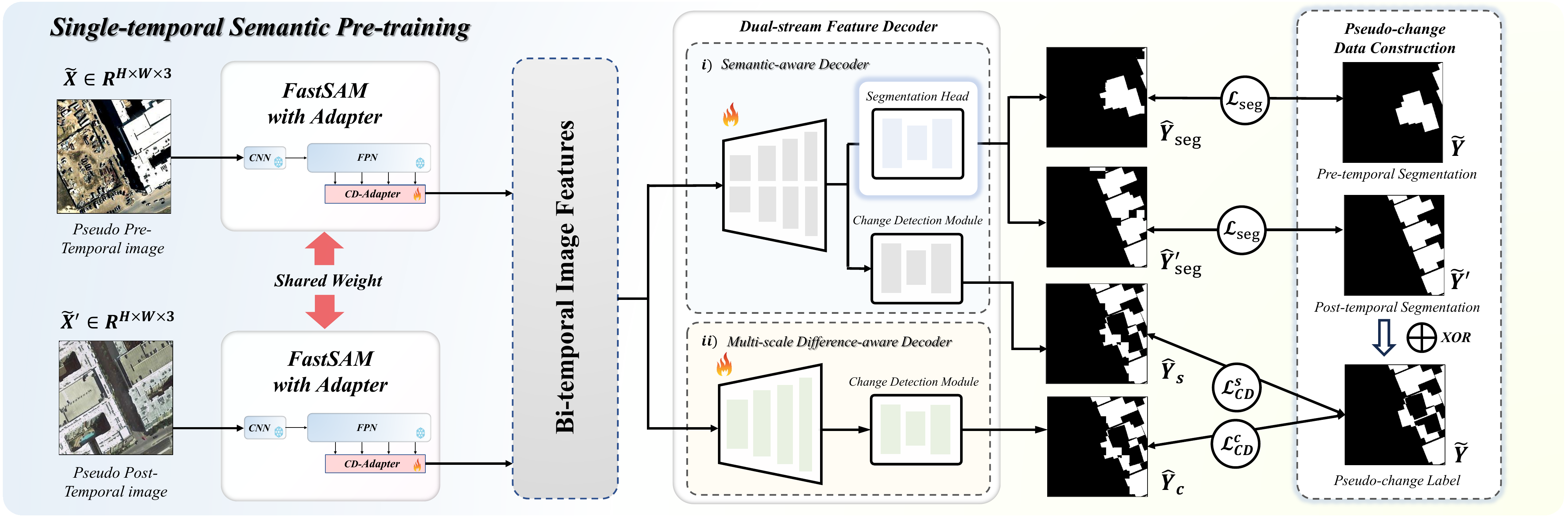}
            \end{center}
            \vspace{0.4cm}
            \caption{\justifying The pipeline of single-temporal semantic pre-training. We first construct pseudo-change data from single-temporal semantic segmentation datasets, providing a large-scale dataset with full annotations. Then, compared with the raw SA-CDNet, we drop the adaptive fusion module and employ an additional semantic segmentation head for a proxy single-temporal segmentation task. This pre-training strategy enhances the knowledge about remote sensing images and landscapes.}
            \label{fig:pretrain}
            \vspace{-0.2cm}
        \end{figure*}

    \subsection{Single-temporal Semantic Pre-training Strategy}
    \label{stsps}
        Though our proposed SA-CDNet utilizes the visual foundation model and a dual-stream feature decoder to enhance the semantic understanding of the landscapes, we suppose it still under-explores the semantic priors for change detection. The reasons are twofold. First, existing change detection datasets are limited and small in scale, as manually collecting registered bi-temporal images and corresponding changed-region annotations are costly and challenging, let alone high-quality ones. Besides, datasets that contain specific interested changes are also scarce. Second, most datasets only provide binary change maps for training, which fails to provide other useful information. For example, there is no information about the landscape category before and after the changes. The above two facts restrict semantic learning when training on change detection datasets. {\thirdrevision To solve the issue, we introduce a single-temporal semantic pre-training strategy following \cite{zheng2021change} for simplicity, which constructs massive pseudo-change data to offer richer semantic priors.} Here are detailed instructions.
        
        \subsubsection{\textbf{Pseudo-change data construction}}
        \label{pretrain: data}
            To relieve the data scarcity issue, {\thirdrevision we adopt the public remote sensing segmentation dataset to construct large-scale bi-temporal images with pseudo-change annotations for pre-training following \cite{zheng2021change}}. We choose segmentation datasets for two reasons: \romannumber{1}) Owing to lower costs of collecting single-temporal images and semantic masks, the data scale of remote sensing segmentation datasets is larger than that of change detection as illustrated in Tab.~\ref{tab:segdat} and Tab.~\ref{tab:cddat}. \romannumber{2}) The semantic segmentation datasets naturally provide semantic masks of the interested landscapes of each single-temporal image, offering better supervision for learning semantics.
            Specifically, we extend the task of detecting changes between registered bi-temporal images to identifying interested landscapes differences between any two images. Given the semantic segmentation dataset $\mathcal{D}=(X_i, Y_i)_{i=1\sim N}$, we randomly sample two single-temporal image samples $(X_i, Y_i)$ and $(X_j, Y_j)$. As shown in Fig.~\ref{fig:pretrain}, we treat the two images as pseudo-bi-temporal images, where $\tilde{X}_k = X_i$ and $\tilde{X}^\prime_k = X_j$. The change label is computed as the XOR of the segmentation maps, where $\tilde{Y}_k = Y_i \oplus Y_j$. Thus, the newly pseudo-change detection dataset $\tilde{\mathcal{D}}$ can be denoted as{\majorrevision :} 
            \begin{align}
            \label{3}
                \tilde{\mathcal{D}} = &\{(\tilde{X}_k, \tilde{X}^\prime_k, \tilde{Y}_k^{seg}, \tilde{Y^\prime }^{seg}_k, \tilde{Y}_k)_{k=1\sim K}|\tilde{X}_k = X_i, \nonumber\\ & \tilde{X}^\prime_k = X_j, \tilde{Y}_k^{seg}=Y_i, \tilde{Y^\prime }^{seg}_k=Y_j, \tilde{Y}_k = Y_i \oplus Y_j \}, 
            \end{align}
            where $(X_i, Y_i), (X_j, Y_j) \in D$, and $K$ is the number of pseudo-change detection samples. 
            One advantage of the pseudo-change data is that both the change map for bi-temporal images and the semantic map for each single-temporal image are available. Note that we can construct at most $K={2\choose N} \gg N$ samples from $N$ single-scale semantic segmentation samples, which considerably enlarges the available data for pre-training. This strategy also leads to diverse samples.

        \subsubsection{\textbf{Semantic learning pre-training}}
        \label{pretrain: strategy} 
            Pre-training endows the network with basic knowledge about images from extensive data with proxy training objectives. It is widely used to relieve the data scarcity issue of downstream tasks. For example, researchers would use ImageNet pre-trained parameters to initialize the backbone ResNet or ViT~\cite{carion2020end,liu2021swin}. {\thirdrevision Thus, we employ the pseudo-change detection dataset $\tilde{D}$ to pre-train SA-CDNet for better adaptation, which provides richer semantics about remote sensing landscapes.}

            As illustrated in Fig.~\ref{fig:pretrain}, we still freeze FastSAM encoder to keep its knowledge intact and pre-train the rest of modules on the constructed dataset, {\it i.e.}, the adapter and the dual-stream decoder. While the whole network structure is unchanged, we dropped the adaptive fusion module. Moreover, we introduce an extra semantic segmentation head $\mathcal{S}(\cdot)$ as the semantic-aware decoder branch, which is constructed by one \texttt{conv + batch norm + ReLU} block and a \texttt{conv + softmax} layer. {\majorrevision The segmentation head receives fused single-temporal features $D_1$ and $D^\prime_1$ of the two input images from the semantic-aware decoder branch, and predicts their segmentation maps $\tilde{Y}_\text{seg}$ and $\tilde{Y^\prime }_\text{seg}$ individually}, which is formulated as{\majorrevision :}
            \begin{equation}
                \hat{Y}_{seg} = \mathcal{S}(D_1), \quad 
                \hat{Y}^\prime_{seg} = \mathcal{S}(D^\prime_1), 
            \end{equation}
            where $\hat{Y}_{seg}$ and $\hat{Y}^\prime_{seg}$ are predicted semantic maps for each image. We also impose constraints to $\hat{Y}_{seg}$ and $\hat{Y}^\prime_{seg}$ and treat it as a proxy semantic segmentation task for semantic pre-training, thanks to the available single-temporal segmentation masks in the pseudo-change dataset. This design exploits the semantic maps to enhance the semantic priors of the interested landscapes. The evidence is revealed in Sec.~\ref{sec:fs-comparison}.

    \subsection{The Overall Training Strategy for SA-CDNet} 
    \label{strategy}    
        We employ a ``pre-training + fine-tuning" pipeline following common practices~\cite{wang2023change,quan2023unified}. In the pre-training phase, we train SA-CDNet with the additional semantic segmentation head $\mathcal{S}(\cdot)$ on the pseudo-change dataset $\tilde{D}$, to distill knowledge of remote sensing images and landscapes from massive pseudo-bi-temporal images. Then, we fine-tune our model on change detection datasets for the final prediction. Here are more details including the loss functions. 

        \subsubsection{\textbf{Pre-training}} 
        \label{pre-training Stage}
            Given the constructed pseudo-change detection dataset $\tilde{D}$ in Sec.~\ref{pretrain: data}, we freeze the visual foundation model encoder $\mathcal{E}(\cdot)$ and train the change detection adapter $\mathcal{A}(\cdot)$ and the dual-stream feature decoder which includes the semantic-aware decoder branch $\mathcal{M}(\cdot)$ and multi-scale difference-aware decoder branch $\mathcal{N}(\cdot)$. The adaptive fusion module is dropped when pre-training. Additionally, a semantic segmentation head is appended to the semantic-aware decoder $\mathcal{S}(\cdot)$ to predict semantic maps of each single-temporal image. Two training objectives are adopted for pre-training, {\it i.e.}, pseudo-change detection and single-temporal semantic segmentation. For the pseudo-change detection, we compute the binary cross-entropy loss for the predicted change maps from the two decoder branches, formulated as{\majorrevision :} 
            \begin{align}
            \label{eq:cd loss}
                L_{\text{CD}} &= L_{\text{CD}}^s + L_{\text{CD}}^c = \mathcal{L}_{ce}(\hat{Y}_s, \tilde{Y}) + \mathcal{L}_{ce}(\hat{Y}_c, \tilde{Y})
            \end{align}
            where $L_{\text{CD}}^s$ and $L_{\text{CD}}^c$ are the losses of semantic-aware and difference-aware decoder branch, respectively. $\mathcal{L}_\text{ce}(\cdot)$ is the cross-entropy loss function. For the proxy single-temporal semantic segmentation task, the loss is computed as{\majorrevision :} 
            \begin{equation}
            \label{eq:seg loss}
                L_{\text{seg}} = \mathcal{L}_{ce}(\hat{Y}_\text{seg}, \tilde{Y}_\text{seg}) + \mathcal{L}_{ce}(\hat{Y}_\text{seg}^\prime, \tilde{Y^\prime }_\text{seg}),
            \end{equation}
            where $\hat{Y}_\text{seg}$ and $\hat{Y}_\text{seg}^\prime$ are the predicted semantic map of bi-temporal images from the segmentation head. Thus, the total loss of the pre-training phase consists of two parts, which can be expressed as follows:
            \begin{equation}
            \label{eq:pretrain}
                L_{\text{pt}} = L_{\text{CD}} + \lambda \cdot L_{\text{seg}}, 
            \end{equation}
            where $\lambda$ is the coefficient to balance the two pre-training objectives. $\lambda$ is empirically set to $1.0$ in our experiments.

        \subsubsection{\textbf{Fine-tuning}} 
        \label{finetuning Stage}
            While the network learns to predict the changes after the pre-training, this model still performs weakly in the real scenes. Evidence can be found in Tab.~\ref{tab:category}. This is because the pre-training only equips the network with basic understandings of remote sensing images and semantic priors of landscapes from the large-scale pseudo-change data. But there still exists a large gap ({\it e.g.}, seasonal and illumination variations) between the pseudo-change data and real ones. Therefore, it is necessary to adapt such knowledge and priors to the real bi-temporal images and interested changes through fine-tuning on the change detection datasets.
            
            Specifically, we initialize SA-CDNet with those after single-temporal semantic pre-training and further fine-tune the network on change detection datasets. The visual foundation model encoder $\mathcal{E}(\cdot)$ is still frozen. Since existing change detection datasets only provide bi-temporal images and difference maps, the segmentation head $\mathcal{S}(\cdot)$ is dropped during fine-tuning. We only fine-tune the change detection adapter $\mathcal{A}(\cdot)$, the semantic-aware decoder branch $\mathcal{M}(\cdot)$ and the multi-scale difference-aware decoder branch $\mathcal{N}(\cdot)$ of the dual-stream feature decoder, and the adaptive fusion module. The overall loss for fine-tuning only contains the change detection loss from the dual-stream feature decoder and the adaptive fusion module, which is computed as{\majorrevision:} 
            \begin{align}
            \label{eq:finetune}
                L_{\text{ft}} &= L_{\text{CD}}^s + L_{\text{CD}}^c + L_{\text{CD}}^f \nonumber \\
                              &= \mathcal{L}_{ce}(\hat{Y}_s, Y) + \mathcal{L}_{ce}(\hat{Y}_c, Y) + \mathcal{L}_{ce}(\hat{Y}, Y), 
            \end{align}
            where $\hat{Y}_s$ and $\hat{Y}_c$ are the predictions of the dual-stream feature decoder. $\hat{Y}$ is the final prediction produced by the adaptive fusion module, which is empirically better than $\hat{Y}_s$ and $\hat{Y}_c$.

            \renewcommand\arraystretch{1.3}
            \begin{table*}[ht]
                \begin{center}
                \caption{Detailed information about the semantic segmentation datasets utilized for pre-training.}
                \label{tab:segdat} 
                \begin{tabular}{!{\centering}m{2.5cm}<{\centering}m{1.3cm}<{\centering}m{2.0cm}<{\centering}m{1.2cm}<{\centering}m{1.0cm}<{\centering}p{7.1cm}<{\centering}}
                \Xhline{1.2pt}
                Dataset  & Spatial Resolution & Image Size  & Image Number  & Class Number  &  Class Information   \\ 
                \Xhline{1.2pt}
                AIRS~\cite{AIRS}  & 0.075~m  &  $10,000 \times10,000$  & 1,046   & 1  & Building \\
                INRIA-Building~\cite{INRIABuilding} & 0.3~m  & $5,000\times5,000$  & 180   & 1  & Building \\
                WHU-Building~\cite{WHUCD}  & 0.3~m  & $512 \times 512$   & 8,189  & 1  & Building   \\
                \hline
                DLCCC~\cite{DLCCC}  & 0.5~m  & $2,448\times 2,448$  & 803  & 7  & Agriculture, barren, forest, rangeland, urban, unknown, water   \\
                LoveDA~\cite{LoveDA} & 0.3~m & $1,024\times 1,024$   & 5,987  & 7  &  Agriculture, barren, building, background, forest, road, water  \\  
                \Xhline{1.2pt}
                \end{tabular}
                \end{center}
                \vspace{-0.2cm}
            \end{table*}
        
            \renewcommand\arraystretch{1.3}
            \begin{table*}[ht]
                \begin{center}
                \caption{Detailed information about change detection datasets used for experiments.}
                \label{tab:cddat}
                \begin{tabular}{!{\centering}m{3.4cm}<{\centering}m{1.5cm}<{\centering}m{2.3cm}<{\centering}m{1.4cm}<{\centering}<{\centering}m{2.8cm}<{\centering}m{2.0cm}<{\centering}m{1.5cm}<{\centering}}
                \Xhline{1.2pt}
                Dataset & Spatial Resolution  & Image Size  & Image Number & Train$/$Val$/$Test  & Cropped Size &  Classes  \\  
                \Xhline{1.2pt}
                LEVIR-CD~\cite{leviercd}  & 0.5~m  & $1,024\times 1,024$   & 637  & $445:64:128$  & $256\times 256$ & Building  \\
                LEVIR-CD+~\cite{leviercd}  & 0.5~m  & $1,024\times 1,024$   & 985 & $637:0:384$ & $256\times 256$ & Building   \\
                S2Looking~\cite{s2looking} & 0.5$\sim$0.8~m  & $1,024\times 1,024$   &  5,000 & $3,500:500:1,000$ & $256\times 256$ & Building \\
                WHU-CD~\cite{WHUCD} & 0.075~m  & $32,507\times 15,354$ & 1 & $-$ & $256\times 256$  & Building  \\
                \hline
                WHU-Cul~\cite{whucultivate} & 1$\sim$2~m  & $512\times 512$ &3,194 & $2,694:0:500$ & $-$ & Farmland \\ 
                \Xhline{1.2pt}
                \end{tabular}
                \end{center}
                \vspace{-0.4cm}
            \end{table*}
    
\section{Experiments}
\label{Experimental results}

    \subsection{Experimental Setup} 
        \subsubsection{\textbf{Datasets and evaluation metrics}}
        \label{sec:dataset}
            We conduct experiments on two kinds of datasets: {\bf \romannumber{1})} {\it Remote sensing semantic segmentation datasets}. We utilize three single-class segmentation datasets, ({\it i.e.}, AIRS~\cite{AIRS}, INRIA-Building~\cite{INRIABuilding}, and WHU-Building~\cite{WHUCD}, for buildings) and two multi-class semantic segmentation datasets ({\it i.e.}, DLCCC~\cite{DLCCC} and LoveDA~\cite{LoveDA}, for 7-class landscapes within agriculture, roads, water, {\it etc.}) for pre-training. These datasets are used to construct pseudo-change data during the pre-training phase. {\bf \romannumber{2})} {\it Change detection datasets}. We evaluate our method on four building change datasets ({\it i.e.}, LEVIR-CD~\cite{leviercd}, LEVIR-CD+~\cite{leviercd}, S2Looking~\cite{s2looking}, WHU-CD~\cite{WHUCD}) and one farmland change dataset ({\it i.e.}, the WHU Cultivated Land dataset, namely WHU-Cul~\cite{whucultivate}). Details about each dataset refer to Tab.~\ref{tab:segdat} and  Tab.~\ref{tab:cddat}. Following previous works~\cite{snunet,bit,samcd}, we report the precision, recall, F1-score, and {\majorrevision{Intersection over Union~(IoU)}} as evaluation metrics for all experiments.

        \subsubsection{\textbf{Implementation details}}
        \label{sec:implementation}
            We conduct all experiments with PyTorch on $8\times$ 32GB NVIDIA Tesla V100 GPUs. We employ SGD optimizer with a weight decay of 0.0005 and a momentum of 0.9. The batch size is 32 and the learning rate decayed following an exponential schedule. When pre-training, we crop all images into $512\times$512 without overlap and randomly pick samples from the cropped segmentation dataset to generate the pseudo-change data online, containing 9000 random samples in one epoch. We pre-train the model for 200 epochs with an initial learning rate of $0.1$. During fine-tuning, we fine-tune on the change detection dataset for 50 epochs with an initial learning rate of 0.01. For fair comparison with previous methods, we crop images into a unified size as listed in Tab.~\ref{tab:cddat}.
            To enhance the robustness of the model, we employ data augmentation strategies including random flips, random rotation, random added noise, and normalization. 

    \subsection{Ablation Study}
    \label{sec:ablation}
        We conduct ablation studies to: {\bf \romannumber{1})} verify the effectiveness of proposed modules, and {\bf \romannumber{2})} discuss different settings for pre-training to obtain the optimal one. 

        \subsubsection{\textbf{Ablations for model design}} We first compare the performance of feature encoders from different visual foundation models, then ablate the change detection adapter and each of the two branches in the dual-stream feature decoder. We also discuss different fusion strategies for the adaptive fusion module. All experiments were conducted on WHU-CD for consistent evaluation. Note that we did not use the proposed pre-training strategy in this section. 
        
        {\bf Comparison of different visual foundation model encoders.} We select SAM and its three variants as the candidates to compare their performance and efficiency for encoding features. We utilize different-sized versions of the foundation model, including SAM-b~({\it basic}), SAM-l~({\it large}), SAM-h~({\it huge}), EfficientSAM-t~({\it lightweight}), EfficientSAM-s~({\it standard}), and FastSAM-s~({\it standard}), FastSAM-x~({\it extended}). Considering the distinct model structures, we only utilize the encoder of these models to extract bi-temporal image features. Then we feed these features to a change detection module consisting of one \texttt{conv + sigmoid} layer and trained it on WHU-CD to produce the change map. The results are listed in Tab.~\ref{tab:backbone}, where the SAM-h was excluded due to the computation limitation. 

        As shown in Tab.~\ref{tab:backbone}, while the SAM-series encoders achieved the best F1-score, {\it i.e.}, SAM-b of $77.17\%$ and SAM-l of $81.81\%$, they also had the largest model size and computation costs. These results reflected that SAM could provide powerful feature representations, but its heavy computation costs somehow limited its application to downstream tasks. In contrast, FastSAM-series encoders exhibit the best FLOPs with the third-highest F1-score as $74.60\%$, which is far more efficient than SAM with acceptable performance. Therefore, we choose FastSAM-x as our feature encoder in all experiments for a trade-off between performance and efficiency. 

        {\bf Effects of each modules.} We study the effectiveness of the proposed change detection adapter $\mathcal{A}(\cdot)$, the semantic-aware decoder branch $\mathcal{M}(\cdot)$ and multi-scale difference-aware decoder branch $\mathcal{N}(\cdot)$. The results are reported in Tab.~\ref{tab:modules}. 
            
        All modules contribute to the final performance and complement each other. The change detection adapter $\mathcal{A}(\cdot)$ improve the precision and recall by a large margin of $23.86\%$ and $4.11\%$, resulting in an improvement of $13.52\%$ F1-score. This fact confirms the gap between natural images and remote sensing images, exhibiting the necessity of the adapter. As for the dual-stream feature decoder, when the multi-scale difference-aware decoder bring an increment of $3.92\%$ F1-score, our semantic-aware decoder further improve it by $0.87\%$. The results show that the multi-scale difference features are vital to change detection, and incorporating semantic understandings into the network brings further advantages. Our method achieve the optimal performance by combining all modules with a $92.91\%$ F1-score, surpassing the baseline by $18.31\%$.

        \renewcommand\arraystretch{1.3}
        \begin{table}[t]
            \begin{center}
            \caption{Comparison of performance and efficiency of different vision foundation model encoders, where \textbf{bold} for the optimal result and \underline{underline} for the second-best one.}
            \label{tab:backbone} 
            \scalebox{0.93}{
            \begin{tabular}{m{1.8cm}<{\centering}|m{1.1cm}<{\centering}m{1.2cm}<{\centering}|m{0.5cm}<{\centering}m{0.5cm}<{\centering}m{0.5cm}<{\centering}m{0.7cm}}
            \Xhline{1.2pt}
            \multirow{2}{*}{Model} & \multicolumn{2}{c}{Computation Complexity} & \multicolumn{4}{c}{Metric} \\ 
            \cline{2-7}
            ~ & Params(Mb)     & FLOPs(G)      & P(\%)    & R(\%)     & F1(\%)    & {\majorrevision{IoU(\%)}}      \\ 
            \Xhline{1.2pt}
            MobileSAM                         & \bf 6.33              & 75.93             & 42.41       & 71.38       & 53.22    & {\majorrevision{36.26}}\\
            \hline
            EfficientSAM-t                      & \underline{6.72}             & 52.52             & 75.65       & 70.18       & 72.81       & {\majorrevision{57.25}}\\
            EfficientSAM-s                      & 22.87             & 184.74             & 76.84       & 70.74       & 73.66       & {\majorrevision{58.30}}\\
            \hline
            FastSAM-s                         & 11.22             & \bf 1.85              & 75.21       & 63.19       & 68.68       & {\majorrevision{52.30}}\\
            FastSAM-x                         & 68.23             & \underline{6.32}              & 71.91       & \underline{77.49}       & 74.60      & {\majorrevision{59.49}} \\
            \hline
            SAM-b                           & 87.03             & 740.45             & \textbf{92.47}   & 66.21       & \underline{77.17}       & {\majorrevision{\underline{62.83}}} \\
            SAM-l                           & 304.54             & 2626.37            & \underline{85.93}       & \textbf{78.06}   & \textbf{81.81}  & \majorrevision{\textbf{69.22}}  \\
            SAM-h      & 632.18        & 5470.24        &$-$      &$-$        &$-$   & {\majorrevision{$-$}}    \\  
            \Xhline{1.2pt}
            \end{tabular}}
            \end{center}
            \vspace{-0.2cm}
        \end{table}

        \renewcommand\arraystretch{1.2}
        \begin{table}[t]
            \begin{center}
            \caption{Effects of each module. $\mathcal{A}(\cdot)$ denotes the adapter, $\mathcal{M}(\cdot)$ denotes the semantic-aware decoder, and $\mathcal{N}(\cdot)$ denotes the multi-scale difference-aware decoder.}
            \label{tab:modules}
            \begin{tabular}{m{0.6cm}<{\centering}m{0.6cm}<{\centering}m{0.6cm}<{\centering}|m{0.9cm}<{\centering}m{0.9cm}<{\centering}m{0.9cm}<{\centering}m{0.9cm}}
            \Xhline{1.2pt}
            \multicolumn{3}{c}{Module} & \multicolumn{4}{c}{Metric} \\
            \Xhline{0.5pt}
            $\mathcal{A}(\cdot)$   & $\mathcal{M}(\cdot)$ & $\mathcal{N}(\cdot)$ & P(\%)     & R(\%)     & F1(\%)   & {\majorrevision{IoU(\%)}}  \\ 
            \Xhline{1.2pt}
            &   &   & 71.91     & 77.49     & 74.60    & {\majorrevision{59.49}} \\
            \hline
            \checkmark  &   &    & \textbf{95.77}     & 81.60     & 88.12  & {\majorrevision{78.76}}   \\ 
            \checkmark     & \checkmark     &     & 93.60     & 88.55     & 91.00     & {\majorrevision{83.49}}\\
            \checkmark     &      & \checkmark    & 94.33    & 89.86    & 92.04   & {\majorrevision{85.25}}  \\
            \hline
            \checkmark    & \checkmark   & \checkmark   & 94.66 & \textbf{91.22} & \textbf{92.91}  & {\majorrevision{\textbf{86.76}}}\\
            \Xhline{1.2pt}
            \end{tabular}
            \end{center}
            \vspace{-0.6cm}
        \end{table}

        \renewcommand\arraystretch{1.2}
        \begin{table}[t]
            \begin{center}
            \caption{Comparison of different fusion strategies.}
            \label{tab:fusion}
            \begin{tabular}{m{2.4cm}<{\centering}|m{0.9cm}<{\centering}m{0.9cm}<{\centering}m{0.9cm}<{\centering}m{0.9cm}}
            \Xhline{1.2pt} 
            Fusion Strategy    & P(\%) & R(\%) & F1(\%)    & {\majorrevision{IoU(\%)}}\\ 
            \Xhline{1.2pt}
            Max               & 93.08 & \bf 91.82 & 92.44 & {\majorrevision{85.94}}\\
            Mean              & \bf 95.27 & 90.01 & 92.57 & {\majorrevision{86.17}}\\
            Learnable Weight  & 94.66 & 91.22 & \bf 92.91 & {\majorrevision{\textbf{86.76}}}\\
            \Xhline{1.2pt}
            \end{tabular}
            \end{center}
            \vspace{-0.2cm}
        \end{table}

        \renewcommand\arraystretch{1.2}
        \begin{table}[t]
            \begin{center}
            \caption{Effects of the semantic segmentation head $\mathcal{S}(\cdot)$ on the pre-training phase.}
            \label{tab:semanticaware}
            \begin{tabular}{m{1.9cm}<{\centering}|m{1.1cm}<{\centering}|m{0.6cm}<{\centering}m{0.6cm}<{\centering}m{0.6cm}<{\centering}m{0.8cm}}
            \Xhline{1.2pt}
            Dataset  & $\mathcal{S}(\cdot)$  & P(\%)  & R(\%)  & F1(\%) & {\majorrevision{IoU(\%)}} \\ 
            \Xhline{1.2pt}
            & ×           & 48.75    & 92.18         & 63.77   & {\majorrevision{46.81}}      \\
            \multirow{-2}{*}{AIRS}   & \checkmark   & \bf 50.88   & \bf 92.94    & \bf 65.76  & {\majorrevision{\textbf{48.99}}}       \\
            \hline
                         & ×    & \bf 47.06    & 86.83    & 61.04    & {\majorrevision{43.93}}     \\
            \multirow{-2}{*}{INRIA-Building} & \checkmark   & 46.84    & \bf 93.35     & \bf 62.38    & {\majorrevision{\textbf{45.33}}}   \\
            \hline
                         & ×           & 57.82    & 91.75    & 70.93   & {\majorrevision{54.95}}\\
            \multirow{-2}{*}{WHU-Building}  & \checkmark  & \bf 58.82   & \bf 93.21  & \bf 72.12 & {\majorrevision{\textbf{56.40}}}     \\  
            \Xhline{1.2pt}
            \end{tabular}
            \end{center}
            \vspace{-0.6cm}
        \end{table}

        \renewcommand\arraystretch{1.1}
        \begin{table}[t]
            \begin{center}
            \caption{Effects of single-class and multi-class segmentation datasets on the pre-training.}
            \label{tab:category}
            \begin{tabular}{m{2cm}<{\centering}|m{1.0cm}<{\centering}m{1.0cm}<{\centering}m{1.0cm}<{\centering}m{1.0cm}}
            \Xhline{1.2pt} 
            Dataset    & P(\%) & R(\%) & F1(\%)  & {\majorrevision{IoU(\%)}} \\ 
            \Xhline{1.2pt}
            AIRS      & 50.88 & 92.94 & 65.76 & {\majorrevision{48.99}}\\
            INRIA-Building & 46.84 & \bf 93.35 & 62.38 & {\majorrevision{45.33}}\\
            WHU-Building  & \bf 58.82 & 93.21 & \bf 72.12 & {\majorrevision{\textbf{56.40}}}\\
            \hline
            DLCCC     & 24.69 & 89.61 & 38.71 & {\majorrevision{24.00}}\\
            LoveDA     & 28.12 & 89.40 & 42.78 & {\majorrevision{27.21}}\\
            \Xhline{1.2pt}
            \end{tabular}
            \end{center}
            \vspace{-0.2cm}
        \end{table}

        \renewcommand\arraystretch{1.1}
        \begin{table}[t]
            \begin{center}
            \caption{Comparison of single-class and multi-class semantic segmentation annotations for the pre-training.}
            \label{tab:samecategory}
            \begin{tabular}{m{2.5cm}<{\centering}|m{0.9cm}<{\centering}m{0.9cm}<{\centering}m{0.9cm}<{\centering}m{0.9cm}}
            \Xhline{1.2pt}
            Dataset      & P(\%) & R(\%) & F1(\%) & {\majorrevision{IoU(\%)}}\\ 
            \Xhline{1.2pt}
            DLCCC       & \bf 70.66 & 28.97 & 41.09 & {\majorrevision{25.86}}\\
            DLCCC-Cultivation & 65.10 & \bf 42.46 & \bf 51.50 & {\majorrevision{\textbf{34.68}}}\\
            \hline
            LoveDA       & \bf 65.22 & 30.76 & 41.80 & {\majorrevision{26.42}}\\
            LoveDA-Cultivation & 65.07 & \bf 44.22 & \bf 52.66 & {\majorrevision{\textbf{35.74}}}\\     
            \Xhline{1.2pt}
            \end{tabular}
            \end{center}
            \vspace{-0.4cm}
        \end{table}

        \renewcommand\arraystretch{1.2}
        \begin{table}[t]
            \begin{center}
            \caption{Effects of different combinations for pre-training.}
            \label{tab:building-combination}
            \scalebox{0.92}{
            \begin{tabular}{m{0.55cm}<{\centering}m{1.86cm}<{\centering}m{1.75cm}<{\centering}|m{0.5cm}<{\centering}m{0.5cm}<{\centering}m{0.5cm}<{\centering}m{0.65cm}}
            \Xhline{1.2pt} 
            \multicolumn{3}{c}{Dataset} & \multicolumn{4}{c}{Metric} \\
            \Xhline{1.2pt}
            AIRS & INRIA-Building  & WHU-Building  & P(\%)   & R(\%)  & F1(\%)  & {\majorrevision{IoU(\%)}} \\ 
            \Xhline{1.2pt}
            \checkmark     &           &          & 50.88       & 92.94       & 65.76   & {\majorrevision{48.99}}    \\
            & \checkmark          &          & 46.84       & 93.35       & 62.38      & {\majorrevision{45.33}} \\
            &           & \checkmark         & 58.82       & 93.21       & 72.12      & {\majorrevision{56.40}} \\
            \hline
            \checkmark     & \checkmark          &          & 53.74       & \textbf{94.16}   & 68.42    & {\majorrevision{52.00}}   \\
            \checkmark     &           & \checkmark         & 64.57       & 91.30       & 75.64   & {\majorrevision{60.82}}    \\
            & \checkmark          & \checkmark         & 62.10       & 93.72       & 73.70     & {\majorrevision{58.35}}  \\
            \hline
            \checkmark     & \checkmark          & \checkmark         & \textbf{65.48}   & 92.41       & \textbf{76.65}  & {\majorrevision \textbf{62.14}} \\ 
            \hline
            \hline
            \end{tabular}}

            \scalebox{0.92}{
            \begin{tabular}{m{2.29cm}<{\centering}m{2.33cm}<{\centering}|m{0.5cm}<{\centering}m{0.5cm}<{\centering}m{0.5cm}<{\centering}m{0.65cm}} 
            DLCCC-Cultivation & LoveDA-Cultivation & P(\%)  & R(\%)   & F1(\%)  & {\majorrevision{IoU(\%)}}\\ 
            \Xhline{1.2pt}
            \checkmark  &   & \textbf{65.10} & 42.46  & 51.50    & {\majorrevision{34.68}} \\
            & \checkmark   & 65.07     & 44.22     & 52.66   & {\majorrevision{35.74}}\\
            \hline
            \checkmark & \checkmark  & 59.87 & \textbf{51.00} & \textbf{55.08} & {\majorrevision \textbf{38.01}}\\
            \Xhline{1.2pt}
            \end{tabular}}
            \end{center}
            \vspace{-0.6cm}
        \end{table}

        {\bf Comparison of different strategies for adaptive fusion module.} Inspired by the ensemble learning, we explore to combine predictions from the two decoder branches with varied strategies. We test three strategies, including computing the max and mean value, and using a learnable weight $\sigma(\omega)$. As listed in Tab.~\ref{tab:fusion}, max fusion benefited the recall, while mean fusion benefited the precision. Fusion with a learnable weight obtained a trade-off with the best $92.91\%$ F1-score. 

        \subsubsection{\textbf{Ablations for pre-training strategy}} We first validate the effectiveness of semantic segmentation head, then explore different settings and combinations of remote sensing semantic segmentation datasets, and analyze their impacts on pre-training. Here we train the network on the pseudo-change dataset and reported its performance on WHU-CD for consistent comparison. We suppose that higher performance on the unseen WHU-CD dataset indicates better knowledge of remote sensing images after pre-training.

        {\bf Effects of the semantic segmentation head $\mathcal{S}(\cdot)$.} We ablate the semantic segmentation head on three buildings change detection datasets as presented in Tab.~\ref{tab:semanticaware}. Equipped with the proposed segmentation head $\mathcal{S}(\cdot)$, the performance increased consistently on all datasets, where the F1-score increased $1.99\%$ on AIRS, $1.34\%$ on INRIA-Building, and $1.09\%$ on WHU-Building. This indicates that the segmentation head $\mathcal{S}(\cdot)$ encourages models to learn semantic priors of the landscapes.
        
        {\bf Comparison of single-class and multi-class segmentation datasets for pre-training.} We employ two kinds of remote sensing semantic segmentation datasets for pre-training: {\bf \romannumber{1})} three single-class segmentation datasets for buildings ({\it i.e.}, AIRS, INRIA-Building, and WHU-Building), and {\bf \romannumber{2})} two multi-class semantic datasets ({\it i.e.}, DLCCC and LoveDA) for buildings, roads, water, {\it etc}. Since the downstream change detection dataset ({\it i.e.}, WHU-CD) only concerns building changes, we discuss the effects of different kinds of datasets on the pre-training. We train the network on each dataset and compared their performance in Tab.~\ref{tab:category}. For fairness, we kept the same number of training samples for all experiments. 

        As illustrated in Tab.~\ref{tab:category}, the model trained on WHU-Building achieved the best performance due to the smallest appearance gap to WHU-CD. Moreover, the models trained on the single-class segmentation datasets outperformed those trained on the multi-class segmentation datasets by a large margin. We suppose the models trained on the multi-class segmentation datasets were confused by different landscape types, resulting in comparable recall but much lower precision than those trained on the single-class datasets. 

        To validate this assumption and avoid potential biases of images, we filter irrelevant categories and only keep the semantic annotations of the same category of downstream tasks. We take farmland changes for example and construct an alternative dataset by filtering annotations except farmland, namely DLCCC-Cultivation and LoveDA-Cultivation. This ensures that only the category number was different between the two datasets. We train the model with the above datasets and report the performance on WHU-Cul in Tab.~\ref{tab:samecategory}. Both models trained with single-class annotations outperform those trained with multi-class annotations, with an obvious increment of about $10\%$ F1-score under both settings. Therefore, it is better to use the segmentation datasets of the same category as the downstream change detection task for pre-training. Including annotations of other landscape categories would disturb the representations of changes of interest. 

        \makeatletter
            \newcommand{\ssymbol}[1]{^{\@fnsymbol{#1}}}
        \makeatother

        \renewcommand\arraystretch{1.4}
        \newcolumntype{C}[1]{>{\centering\arraybackslash}p{#1}}
        \begin{table*}[t]
            \begin{center}
            \caption{Comparisons with existing SOTA methods on LEVIR-CD, LEVIR-CD+, S2Looking, WHU-CD, and WHU-Cul. }
            \label{tab:building}
            \scalebox{0.75}{
            \begin{tabular}{m{2.35cm}<{\centering}|m{1.4cm}<{\centering}|m{0.5cm}<{\centering}m{0.5cm}<{\centering}m{0.5cm}<{\centering}m{0.69cm}<{\centering}|m{0.5cm}<{\centering}m{0.5cm}<{\centering}m{0.5cm}<{\centering}m{0.69cm}<{\centering}|m{0.5cm}<{\centering}m{0.5cm}<{\centering}m{0.5cm}<{\centering}m{0.69cm}<{\centering}|m{0.5cm}<{\centering}m{0.5cm}<{\centering}m{0.5cm}<{\centering}m{0.69cm}<{\centering}|m{0.5cm}<{\centering}m{0.5cm}<{\centering}m{0.5cm}<{\centering}m{0.69cm}<{\centering}}
            \Xhline{1.2pt}
            ~ & ~ & \multicolumn{4}{c}{LEVIR-CD} & \multicolumn{4}{c}{LEVIR-CD+} & \multicolumn{4}{c}{S2Looking} & \multicolumn{4}{c}{WHU-CD} & \multicolumn{4}{c}{WHU-Cul} \\ 
            \cline{3-22}
            \multirow{-2}{*}{Method}   & {\minorrevision \multirow{-2}{*}{Backbone}}    & P(\%)      & R(\%)      & F1(\%)  & {\majorrevision{IoU(\%)}}   & P(\%)      & R(\%)      & F1(\%)  & {\majorrevision{IoU(\%)}}   & P(\%)      & R(\%)      & F1(\%) & {\majorrevision{IoU(\%)}}  & P(\%)      & R(\%)      & F1(\%)  & {\majorrevision{IoU(\%)}}  & P(\%)      & R(\%)      & F1(\%)  & {\majorrevision{IoU(\%)}}\\ 
            \Xhline{1.2pt}
            FC-EF~\cite{daudt2018fully}    & {\minorrevision CNN}   & 86.91      & 80.17      & 83.40   & {\majorrevision{71.53}}   & 76.49      & 76.32      & 76.41  & {\majorrevision{61.83}}     & \underline{81.36}      & 8.95       & 7.65  & {\majorrevision{3.98}}  & 80.75      & 67.29      & 73.40   & {\majorrevision{57.98}}  & 60.29        & 62.98        & 61.61  & {\majorrevision{44.52}} \\
            FC-Siam-conc~\cite{daudt2018fully}  & {\minorrevision CNN} & 91.99  & 76.77  & 83.69  & {\majorrevision{71.95}}  & 81.12  & 77.16      & 79.09  & {\majorrevision{65.41}}   & 68.27   & 18.52   & 13.54  & {\majorrevision{7.26}}  & 54.20  & 81.34  & 65.05  & {\majorrevision{48.20}}  & 62.51  & 65.24  & 63.85  & {\majorrevision{46.90}} \\
            FC-Siam-diff~\cite{daudt2018fully}  & {\minorrevision CNN}  & 89.53   & 83.31   & 86.31   & {\majorrevision{75.92}}   & 80.88      & 77.65  & 79.23  & {\majorrevision{65.60}}  & \textbf{83.49}  & 32.32   & 46.60 & {\majorrevision{30.38}}  & 48.84   & 88.96  & 63.06  & {\majorrevision{46.05}}  & 64.81   & 56.42  & 60.33  & {\majorrevision{43.19}}  \\
            STANet~\cite{su2022stanet}  & {\minorrevision ResNet-18}  & 83.81   & \underline{91.00} & 87.26  & {\majorrevision{77.40}}    & 74.20      & \underline{83.90} & 78.80  & {\majorrevision{65.02}}   & 36.40   & 68.20   & 47.40 & {\majorrevision{31.06}}   & 77.40  & 90.30  & 83.35  & {\majorrevision{71.45}}   & 62.75  & 69.47    & 65.94  & {\majorrevision{49.19}} \\
            DASNet~\cite{chen2020dasnet}  & {\minorrevision ResNet-50} & 80.76  & 79.53  & 79.91  & {\majorrevision{66.54}}  & 77.51  & 78.03  & 77.77  & {\majorrevision{63.63}}  & 45.06 & 48.71  & 47.29  & {\majorrevision{30.97}}  & 83.77 & 91.02   & 87.24  & {\majorrevision{77.37}}  & 60.58   & \textbf{77.00}   & 67.81 & {\majorrevision{51.30}}\\
            SNUNet~\cite{snunet}  & {\minorrevision UNet}  & 89.18      & 87.17      & 88.16  & {\majorrevision{78.83}}    & 78.90      & 78.23      & 78.56   & {\majorrevision{64.69}}   & 45.25      & 50.60      & 47.78  & {\majorrevision{31.39}}   & 91.28      & 87.25      & 89.22  & {\majorrevision{80.54}}   & 68.88        & 72.09        & 70.45  & {\majorrevision{54.38}} \\
            BIT~\cite{bit}    & {\minorrevision ResNet-18}     & 92.57      & 87.65      & 90.04  & {\majorrevision{81.88}}    & 82.37      & 79.73      & 81.03   & {\majorrevision{68.11}}   & 70.26      & 56.53      & 62.65  & {\majorrevision{45.61}}   & 91.56      & 87.84      & 89.66  & {\majorrevision{81.26}}   & 70.84        & 70.11        & 70.48 & {\majorrevision{54.42}} \\
            {\majorrevision{MFPNet~\cite{rs13153053}}}   & {\minorrevision ResNet-50}  & {\majorrevision \textbf{92.85}}   & {\majorrevision 89.41}   & {\majorrevision 91.10}  & {\majorrevision 83.66}    & {\majorrevision \textbf{87.00}}   & {\majorrevision 81.42}    & {\majorrevision \underline{84.12}}   & {\majorrevision \underline{72.59}}   & {\majorrevision 71.55}   & {\majorrevision 54.07}   & {\majorrevision 61.60}  & {\majorrevision 44.50}   & {\majorrevision 86.39}    & {\majorrevision 85.47}    & {\majorrevision 85.92}  & {\majorrevision 75.32}   & {\majorrevision 72.78}     & {\majorrevision 70.87}     & {\majorrevision 71.81}  & {\majorrevision 56.02}  \\ 
            FTN~\cite{yan2022fully}    & {\minorrevision Swin Trans}     & \underline{92.71}      & 89.37      & 91.01    & {\majorrevision{83.50}}  & 80.91      & 83.47      & 82.17   & {\majorrevision{69.74}}   & 61.54      & 60.26      & 63.19  & {\majorrevision{46.19}}   & 94.73      & 89.83      & 92.21  & {\majorrevision{85.55}}   & \underline{75.54}        & 63.20        & 68.82  & {\majorrevision{52.46}}\\
            {\majorrevision{ChangeFormer~\cite{changeformer}}}  & {\minorrevision Transformer}  & {\majorrevision{92.32}}   & {\majorrevision{88.50}}   & {\majorrevision{90.37}}  & {\majorrevision{82.43}}    & {\majorrevision 84.29}   & {\majorrevision 80.81}    & {\majorrevision 82.51}   & {\majorrevision 70.23}   &  {\majorrevision 70.73}  & {\majorrevision 49.25}   & {\majorrevision 58.07}  & {\majorrevision 40.91}   & {\majorrevision 91.76}    & {\majorrevision 84.85}    & {\majorrevision 88.17}  & {\majorrevision 78.85}   & {\majorrevision 68.40}     & {\majorrevision 63.70}     & {\majorrevision 65.97}  & {\majorrevision 49.22}  \\
            VcT~\cite{jiang2023vct}    & {\minorrevision ResNet-18}     & 89.24      & 89.37      & 89.31   & {\majorrevision{80.68}}   & 80.50      & 81.41      & 80.95   & {\majorrevision{68.00}}   & 61.49      & \underline{68.25}  & 62.30  & {\majorrevision{45.24}}   & 93.24      & 76.58      & 84.09   & {\majorrevision{72.55}}  & 65.90        & 66.74        & 66.32  & {\majorrevision{49.61}}\\
            {\majorrevision{AMTNet~\cite{liu2023attention}}}   & {\minorrevision ResNet-50}  & {\majorrevision{91.59}}   & {\majorrevision{89.87}}   & {\majorrevision{90.72}}  & {\majorrevision{83.02}}    & {\majorrevision 83.05}   & {\majorrevision 83.72}    & {\majorrevision 83.38}   & {\majorrevision 71.50}   & {\majorrevision 56.95}   & {\majorrevision \textbf{69.05}}   & {\majorrevision 62.42}  & 45.36   & {\majorrevision{88.90}}    & {\majorrevision{90.21}}    & {\majorrevision{89.55}}  & {\majorrevision{81.08}}   & {\majorrevision 71.04}     & {\majorrevision 65.78}     & {\majorrevision 68.31}  & {\majorrevision 51.87}  \\
            RFL-CDNet~\cite{gan2024rfl}   & {\minorrevision UNet}   & 91.62      & 90.40      & 90.98  & {\majorrevision{83.45}}    & 79.95      & \textbf{84.04}      & 81.94  & {\majorrevision{69.41}}    & 65.72      & 60.82      & 63.17  & {\majorrevision{46.17}}   & 91.33      & \underline{91.46}      & 91.39  & {\majorrevision{84.15}}   & 70.87        & 73.75        & 72.28   & {\majorrevision{56.59}}\\
            SAM-CD~\cite{samcd}    & {\minorrevision FastSAM-x}    & 91.38      & {90.45}      & 90.91  & {\majorrevision{83.33}}    & 79.71      & 81.35      & 81.96   & {\majorrevision{69.43}}   & 74.32      & 55.30      & 63.42  & {\majorrevision{46.43}}   & \textbf{96.87} & 85.67      & 90.92  & {\majorrevision{83.35}}   & 68.80        & 73.12        & 70.89     & 54.91  \\
            \thirdrevision{Meta-CD~\cite{gao2025combining}}    & {\thirdrevision FastSAM-x}    & \thirdrevision{92.51}     & \thirdrevision{88.89}      & \thirdrevision{90.66}  & \thirdrevision{82.91}    & \thirdrevision{80.17}      & \thirdrevision{84.13}      & \thirdrevision{82.10}   & \thirdrevision{69.93}   & \thirdrevision{74.08}      & \thirdrevision{54.00}      & \thirdrevision{62.47}  & \thirdrevision{45.42}   & \thirdrevision{89.00} & \thirdrevision{90.35}      & \thirdrevision{89.67}  & \thirdrevision{81.27}   & \thirdrevision{73.72}        & \thirdrevision{67.31}        & \thirdrevision{70.37}     & \thirdrevision{54.28}  \\
            \hline
            \makecell{Ours ({\it w/o pre-train})}  & {\minorrevision FastSAM-x}  & 92.69      & 89.66      & \underline{91.15}  & {\majorrevision{\underline{83.74}}} & \underline{86.88}      & {79.15}  & 82.83  & {\majorrevision 70.69} & 81.25      & 54.73      & \underline{65.40}  & {\majorrevision{\underline{48.59}}}  & 94.66      & 91.22      & \underline{92.91} & {\majorrevision{\underline{86.76}}}  & 72.65        & \underline{73.93}   & \underline{73.28} & {\majorrevision{\underline{57.83}}}\\
            \makecell{Ours ({\it w pre-train})}  & {\minorrevision FastSAM-x} & {91.77}  & \textbf{91.28}   & \textbf{91.53}  & {\majorrevision{\textbf{84.38}}} & 85.55  & {83.44}  & \textbf{84.43} & {\majorrevision{\textbf{73.06}}} & 81.28      & 56.24      & \textbf{66.48}  & {\majorrevision{\textbf{49.79}}}  & \underline{95.29}      & \textbf{93.67} & \textbf{94.47} & {\majorrevision{\textbf{89.52}}}    & \textbf{77.74}   & 72.82       & \textbf{75.20}  & {\majorrevision{\textbf{60.26}}}  \\
            \hline
            \thirdrevisiont{TTP~\cite{chen2024time}}    & \thirdrevisiont{SAM-l}    & \thirdrevisiont{93.05}     & \thirdrevisiont{91.49}      & \thirdrevisiont{92.26}  & \thirdrevisiont{85.63}    & \thirdrevisiont{85.81}      & \thirdrevisiont{84.36}      & \thirdrevisiont{85.08}   & \thirdrevisiont{74.03}   & \thirdrevisiont{73.51}      & \thirdrevisiont{62.19}      & \thirdrevisiont{67.38}  & \thirdrevisiont{50.80}   & \thirdrevisiont{96.05} & \thirdrevisiont{92.76}      & \thirdrevisiont{94.37}  & \thirdrevisiont{89.34}   & \thirdrevisiont{77.54}        & \thirdrevisiont{73.14}        & \thirdrevisiont{75.27}     & \thirdrevisiont{60.35}  \\
            \thirdrevisiont{SFCD-Net~\cite{zhang2024integrating}}    & {\thirdrevisiont{SAM-l}}    & \thirdrevisiont{92.78}     & \thirdrevisiont{91.59}      & \thirdrevisiont{92.18}  & \thirdrevisiont{85.49}    & \thirdrevisiont{87.25}      & \thirdrevisiont{85.65}      & \thirdrevisiont{86.44}   & \thirdrevisiont{76.12}   & \thirdrevisiont{72.60}      & \thirdrevisiont{61.70}      & \thirdrevisiont{66.71}  & \thirdrevisiont{50.05}   & \thirdrevisiont{95.42} & \thirdrevisiont{93.14}      & \thirdrevisiont{94.27}  & \thirdrevisiont{89.16}   & \thirdrevisiont{71.86}        & \thirdrevisiont{76.46}        & \thirdrevisiont{74.09}     & \thirdrevisiont{58.84}  \\
            \Xhline{1.2pt}
            \end{tabular}}
            \end{center}
            \vspace{-0.4cm}
        \end{table*}
        
        {\bf Effects of varied combinations of segmentation datasets for pre-training.} Considering remote sensing images from the same dataset were captured in close periods on the neighboring regions, they share similar data distribution. This potentially limits the diversity of pseudo-change data. Thus, we use varied combinations to explore the effects of data diversity. For fairness, we set the same number of training samples among different combinations in this section. If there were multiple datasets, the proportion will be $1\colon1$. Tab.~\ref{tab:building-combination} presents the results on WHU-CD and WHU-Cul. The more diverse datasets used, the better performance they would achieve. This observation is consistent among experiments. Thus, the combination of all single-class segmentation datasets of the same category for pre-training yielded the best performance. We use all building segmentation datasets ({\it i.e.}, AIRS, INRIA-Building, and WHU-Building) for detecting building changes and the filtered datasets with only farmland ({\it i.e.}, DLCCC-Cultivation and LoveDA-Cultivation) for farmland changes.

        \renewcommand\arraystretch{1.2}
        \begin{table}
            \begin{center}
            \caption{{\minorrevision Comparisons of F1-score under semi-supervised setup on WHU-CD. $\ssymbol{2}$ means reproduced results.}}
            \label{tab:semi}
            \begin{tabular}{p{2.8cm}<{\centering}|p{0.9cm}<{\centering}p{0.9cm}<{\centering}p{0.9cm}<{\centering}p{0.9cm}<{\centering}}
            \Xhline{1.2pt}
                 & \multicolumn{4}{c}{Proportion of labeled data} \\ \cline{2-5}
            \multirow{-2}{*}{Method} & 5\%       & 10\%      & 20\%      & 40\%      \\ \Xhline{1.2pt}
            SemiCDNet~\cite{peng2020semicdnet}                     & 82.90      & 85.28      & 86.57      & 87.74      \\
            SemiSANet~\cite{sun2022semisanet}                     & 80.95      & 85.32      & 86.99      & 88.36      \\
            MTCNet~\cite{shu2022mtcnet}               & 87.63      &  89.63      & 90.64      & 91.46      \\
            {SemiCD-VL$\ssymbol{2}$}~\cite{semicd_vl}  &  \bf{91.32}   &  \underline{\minorrevision{91.97}}     & \underline{\minorrevision{92.36}}    &  \underline{\minorrevision{92.86}}      \\
            \hline
            Ours ({\it w/o pre-train})        & 87.33      & 89.57      & 91.18      & 92.17   \\
            Ours ({\it w/ pre-train})       & \underline{89.63}  & \textbf{92.77} & \textbf{93.77}  & \textbf{94.13} \\
            \Xhline{1.2pt}
            \end{tabular}
            \end{center}
            \vspace{-0.3cm}
        \end{table}

        \renewcommand\arraystretch{1.2}
        \begin{table}[t]
            \begin{center}
            \caption{Model size and computational complexity. {\majorrevision The F1-score and IoU are reported on {\minorrevision WHU-CD for reference}.}}
            \label{tab:complex} 
            \begin{tabular}{m{2.0cm}<{\centering}|m{0.8cm}<{\centering}m{1.2cm}<{\centering}m{1.0cm}<{\centering}m{0.5cm}<{\centering}m{0.5cm}<{\centering}}
            \Xhline{1.2pt} 
            Method  & Params (Mb) & Trainable Params(Mb) & FLOPs (G) & {\majorrevision F1 (\%)} & {\majorrevision IoU (\%)} \\
            \Xhline{1.2pt}
            STANet~\cite{su2022stanet}  & 16.93   & 16.93  & 6.58 & {\minorrevision 83.35}  & {\minorrevision 71.45}      \\
            DASNet~\cite{chen2020dasnet}  & 48.22   & 48.22  & 25.17    & {\minorrevision 87.24}  & {\minorrevision 77.37}      \\
            SNUNet~\cite{snunet}  & 27.07   & 27.07  & 27.44   & {\minorrevision 89.22}  & {\minorrevision 80.54}        \\
            BIT~\cite{bit}    & \underline{11.47}   & 11.47   & 19.60  & {\minorrevision 89.66}  & {\minorrevision 81.26}       \\
            \minorrevision MFPNet~\cite{rs13153053}    & {\minorrevision 85.97}   & {\minorrevision 85.97}   & {\minorrevision 32.24}  & {\minorrevision 85.92}  & {\minorrevision 75.32}       \\
            \minorrevision FTN~\cite{yan2022fully}    & {\minorrevision 164.45}   & {\minorrevision 164.45}   & {\minorrevision 44.41}  & {\minorrevision \underline{92.21}}  & {\minorrevision \underline{85.55}}       \\
            \minorrevision ChangeFormer~\cite{changeformer}    & {\minorrevision 41.03}   & {\minorrevision 41.03}   & {\minorrevision 50.70}  & {\minorrevision 88.17}  & {\minorrevision 78.85}       \\
            \minorrevision VcT~\cite{jiang2023vct}    & {\minorrevision \bf 3.50}   & {\minorrevision \underline{3.50}}   & {\minorrevision \bf 2.66}  & {\minorrevision 84.09}  & {\minorrevision 72.55}       \\
            \minorrevision AMTNet~\cite{liu2023attention}    & {\minorrevision 24.67}   & {\minorrevision 24.67}   & {\minorrevision \underline{5.39}}  & {\minorrevision 89.55}  & {\minorrevision 81.08}       \\
            RFL-CDNet~\cite{gan2024rfl} & 27.24   & 27.24   & 33.89  & {\minorrevision 91.39}  & {\minorrevision 84.15}       \\
            SAM-CD~\cite{samcd}  & 70.49   & \bf 2.49  & 8.60  & {\minorrevision 90.92}  & {\minorrevision 83.35}       \\
            \thirdrevision{Meta-CD~\cite{gao2025combining}}  & \thirdrevision{81.67}   & \thirdrevision{13.67}  & \thirdrevision{11.93}  & \thirdrevision{89.67}  & \thirdrevision{81.27}       \\
            \hline
            Ours   & 77.08   & 9.08  & 12.03  & {\minorrevision \textbf{94.47}}  & {\minorrevision \textbf{89.52}}        \\
            \hline
            \thirdrevisiont{TTP~\cite{chen2024time}}  & \thirdrevisiont{319.64}   & \thirdrevisiont{11.54}  & \thirdrevisiont{72.63}  & \thirdrevisiont{94.37}  & \thirdrevisiont{89.34}       \\
            \thirdrevisiont{SFCD-Net~\cite{zhang2024integrating}}  & \thirdrevisiont{310.45}   & \thirdrevisiont{14.56}  & \thirdrevisiont{72.22}  & \thirdrevisiont{94.27}  & \thirdrevisiont{89.16} \\    
            \Xhline{1.2pt}
            \end{tabular}
            \end{center}
            \vspace{-0.8cm}
        \end{table}

    \subsection{Comparison with Fully Supervised Methods}
    \label{sec:fs-comparison}
        To demonstrate the superiority of our SA-CDNet, we conduct experiments on several challenging datasets, including four building change datasets ({\it i.e.}, LEVIR-CD, LEVIR-CD+, S2Looking, WHU-CD) and one cultivation change dataset ({\it i.e.}, WHU-Cul). The comparison results are reported in Tab.~\ref{tab:building}, where both of our methods with and without the pre-training are listed for comprehensive comparisons. {\minorrevision Due to the lack of results on some datasets ({\it e.g.,} S2Looking, WHU-Cul), we implemented them with released codes following official instructions and default settings for completeness. We reproduced recent approaches as possible for fair comparison, to avoid potential differences in data splits and preparation. The results of early-published methods are referred to \cite{bit,changeformer,gan2024rfl}.}
        
        {\thirdrevision As shown in Tab.~\ref{tab:building}, our method, which uses a much lighter FastSAM-x~($68.23$M) backbone, achieves competitive performance compared to those using SAM-l~($304.54$M) referring to Tab.~\ref{tab:backbone}. Apart from those methods using SAM-l as the backbone, our method achieves the best F1-score on all five datasets, outperforming the current SOTA methods and concurrent FastSAM-based methods even without the proposed single-temporal semantic pre-training.} Specifically, our method without pre-pretraining surpass the current SOTA method by $0.14\%$ F1-score on LEVIR-CD, $1.98\%$ on S2Looking, $0.7\%$ on WHU-CD, and $1.0\%$ on WHU-Cul. Equipped with the proposed pre-training strategy with the pseudo-change data, our method obtains the best performance, setting new SOTA on five datasets with a $91.53\%$ F1-score on LEVIR-CD, a $84.43\%$ F1-score on LEVIR-CD+, a $66.48\%$ F1-score on S2Looking, a $94.47\%$ F1-score on WHU-CD, and a $75.20\%$ F1-score on LEVIR-CD+, which also surpass the results without pre-training by a large margin. Moreover, our method exhibits more significant increments on more challenging datasets like LEVIR-CD+, S2Looking, and WHU-Cul, which contain complex background and delicate changes. These results prove the advance of SA-CDNet and the effectiveness of single-temporal semantic pre-training, demonstrating the significance of landscape semantic priors in change detection tasks. 

\begin{figure*}[ht!]
            \vspace{-0.4cm}
            \setlength{\abovecaptionskip}{-10pt} 
            \setlength{\belowcaptionskip}{-1pt}
            \begin{center}
            \includegraphics[width=0.94\textwidth]{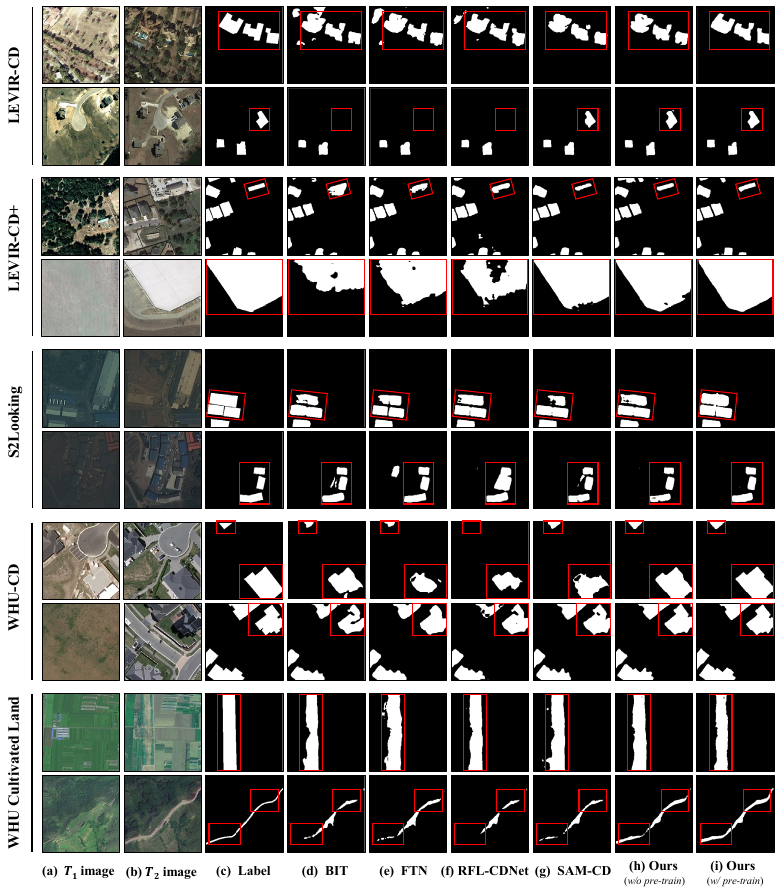}
            \end{center}
            \vspace{0.2cm}
            \caption{Visualized comparison of our method with previous SOTA methods.}
            \label{fig:visualization}
            \vspace{-0.2cm}  
        \end{figure*}

        The visualized comparisons are presented in Fig.~\ref{fig:visualization}. For page limitations, we only show the results of recent advanced methods. As shown in Fig.~\ref{fig:visualization}, our method excels at capturing fine details along boundaries and subtle changes. For example, our method keeps more clear boundaries of the changed regions compared to other methods as illustrated in the $1^{st}, 3^{rd}, 4^{th}, 5^{th}, 7^{th}, 8^{th}$ rows of samples. We attribute these benefits to better semantic understandings of the landscapes, producing more complete and clear regions. This fact proves the effectiveness of SA-CDNet and the semantic pre-training strategy. Incorporating semantics into change detection also helps to avoid false positive and false negative predictions as shown in the $2^{nd}, 6^{th}, 7^{th}$ rows of samples. Moreover, when facing challenging farmland changes with arbitrary shapes in the WHU-Cul, SA-CDNet can produce faithful and robust results with precise boundaries. This fact further verifies the capability of our method of detecting other challenging categories of changes besides the building.       

    \subsection{Comparison with Semi-supervised Methods}
    \label{sec:ss-comparison}
        SA-CDNet with the pre-training strategy also provides a solution when only limited annotated change detection data is available. {\minorrevision We validate this by comparing SA-CDNet with four recent semi-supervised change detection methods~\cite{peng2020semicdnet,sun2022semisanet,shu2022mtcnet,semicd_vl} under the semi-supervised learning setup, as shown in Tab.~\ref{tab:semi}}. To reduce the cost of human annotations, the semi-supervised methods aim to learn from large-scale unlabeled data with the help of limited labeled data. Popular semi-supervised learning techniques utilize consistency regularization and pseudo-labels to mine useful information from large-scale unlabeled data. Here, we test four semi-supervised methods on WHU-CD under the settings of $5\%$, $10\%$, $20\%$, and $40\%$ labels, respectively. All bi-temporal images are accessible during the training if needed. Since our method is not designed for semi-supervised change detection, we only use the labeled images for training and ignore the unlabeled ones.

        {\minorrevision As listed in Tab.~\ref{tab:semi}, without utilizing any unlabeled bi-temporal images, our method without pre-training can achieve competitive results with recent semi-supervised methods, {\it i.e.}, MTCNet~\cite{shu2022mtcnet} and SemiCD-VL~\cite{semicd_vl}. It exhibits the powerful feature representations of foundation models. As the number of labeled data grows, the performance of all methods increases unsurprisingly. Notably, owing to prior knowledge learned from the pre-training, our method with pre-training obtained slightly lower results than the optimal method SemiCD-VL under the $5\%$ setting, and perform best under other settings. These results demonstrate the data efficiency of our method, realizing promising results with limited labeled data.}

    \subsection{Model Size and Computational Complexity}
    \label{sec:computation-cost}

        \begin{table}
            \renewcommand\arraystretch{1.5}
            \begin{center}
            \caption{{\majorrevision Time cost of pre-training and fine-tuning.}}
            \label{tab:timecost}
            \scalebox{0.88}{
            \begin{tabular}{m{1.0cm}<{\centering}|m{2.1cm}<{\centering}|m{0.6cm}<{\centering}m{0.8cm}<{\centering}m{0.8cm}<{\centering}m{0.7cm}<{\centering}m{0.7cm}<{\centering}}
            \Xhline{1.2pt} 
            {\majorrevision Phase}  & {\majorrevision Pre-train}  &  \multicolumn{5}{c}{{\majorrevision Fine-tune}} \\ 
            \hline 
            {\majorrevision Data}   & {\majorrevision Pseudo-change Data}  &  {\majorrevision LEVIR}  &  {\majorrevision LEVIR+}  & {\majorrevision S2-Looking} & {\majorrevision WHU-CD} & {\majorrevision WHU-Cul}  \\
            \Xhline{1.2pt}
            {\majorrevision Time~(h)} & {\majorrevision $\sim$18.5} & {\majorrevision 6.2} & {\majorrevision 9.8} & {\majorrevision 35.3} & {\majorrevision 2.5} & {\majorrevision 2.0}  \\
            \hline
            {\majorrevision $\Delta$F1~(\%)}   & --   & {\majorrevision +0.64}  & {\majorrevision +2.37}  & {\majorrevision +1.2}  & {\majorrevision +2.76}  & {\majorrevision +2.43}        \\
            \Xhline{1.2pt}
            \end{tabular}}
            \end{center}
            \vspace{-0.4cm}
        \end{table}

        Tab.~\ref{tab:complex} presents the model size and computational complexity of our method and some previous SOTA methods. {\thirdrevision Though our method contains $77.08 \text{MB}$ parameters, we highlight that most parameters are frozen, where the frozen FastSAM-x encoder contains about $68\text{MB}$ parameters, and only $9.08\text{MB}$ are trainable. The FLOPs of ours is $12.03\text{G}$, a little higher than that of STANet~($6.58\text{G}$), SAM-CD~($8.60\text{G}$), and Meta-CD~($11.93\text{G}$), but significantly lower than other recent advanced methods. Besides, our method achieved competitive results with $\sim$$4$ times fewer parameters and $\sim$$6$ times fewer FLOPs than SAM-l based TTP and SFCD-Net, showing its efficiency.} Since our method achieves SOTA performance on several change detection benchmarks and surpasses the second-highest one by a large margin, the negligible increment of FLOPs is acceptable. {\majorrevision The time cost of pre-training and fine-tuning is listed in Tab.~\ref{tab:timecost} for reference.}

\section{Conclusion}
\label{Conclusion} 
    To achieve advanced change detection performance, we argue that the network requires to not only perceive the difference between bi-temporal images but also have knowledge of landscapes. {\thirdrevision To this end, this paper proposes a novel SA-CDNet and incorporates a single-temporal semantic pre-training strategy to provide richer semantic priors for change detection.} SA-CDNet mainly consists of a visual foundation model-based feature encoder to inherit knowledge about natural images, an adapter to align features to change detection, and a dual-stream feature decoder to exploit difference-aware and semantic-aware features following the human visual paradigm to distinguish changes. With the help of semantic pre-training, SA-CDNet learns richer priors about the interested landscapes in the remote sensing context from pseudo-change data. We conducted extensive experiments on five benchmarks, where our method achieved the SOTA performance. Detailed ablation studies validated the effectiveness of our designs of both the network structure and the semantic pre-training strategy. 

\bibliographystyle{IEEEtran}
\bibliography{reference} 

\vspace{-15pt}

\begin{IEEEbiography}[{\includegraphics[width=1in,height=1.25in,clip,keepaspectratio]{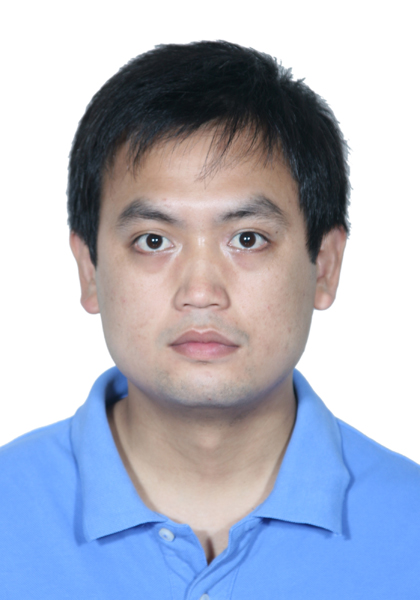}}]{Yuhang Gan} is currently pursuing his Ph.D. degree in Electronic Information Engineering from the School of Computer Science, Wuhan University. His research interests mainly include image processing, computer vision and machine learning. \end{IEEEbiography}

\vspace{-10pt}

\begin{IEEEbiography}[{\includegraphics[width=1in,height=1.25in,clip,keepaspectratio]{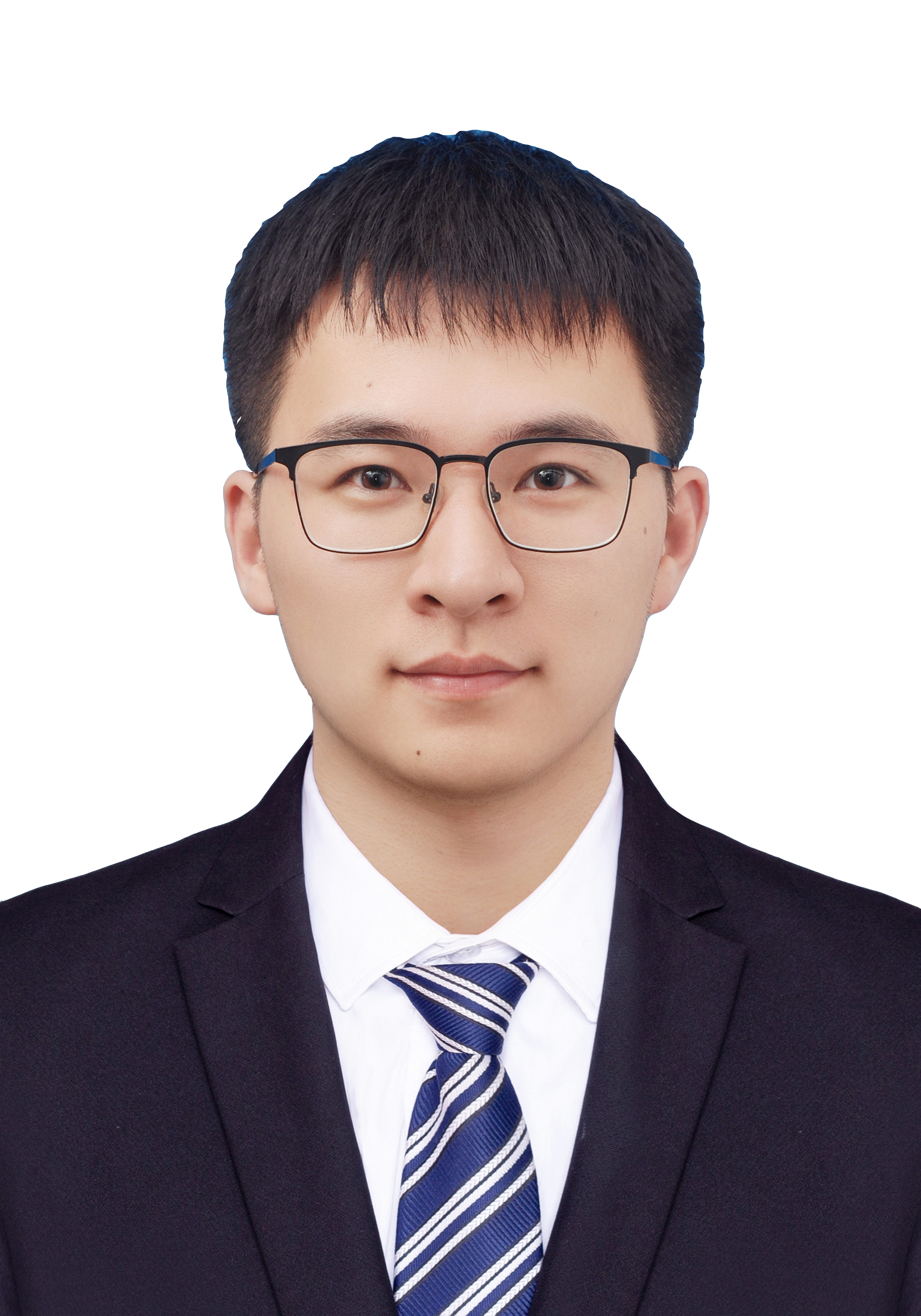}}]{Wenjie Xuan} is currently pursuing his Ph.D. degree in Artificial Intelligence from the School of Computer Science, Wuhan University. His research interests mainly include image processing, computer vision and machine learning. He has published several research papers in \textit{Neural Networks}, \textit{Pattern Recognition}, ACM MM, and ICCV, \textit{etc}.\end{IEEEbiography}

\vspace{-10pt}

\begin{IEEEbiography}[{\includegraphics[width=1in,height=1.25in,clip,keepaspectratio]{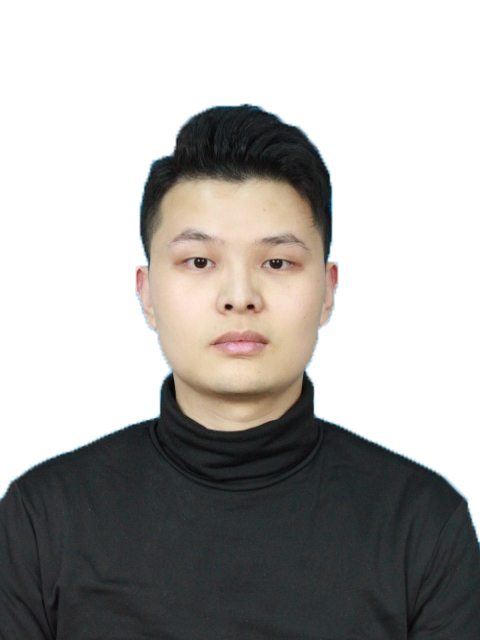}}]{ZhiMing Luo} is currently pursuing his Ph.D. degree at the School of Computer Science, Wuhan University, Wuhan, China. His research interests include remote sensing Image processing and computer vision.  \end{IEEEbiography}

\vspace{-10pt}

\begin{IEEEbiography}[{\includegraphics[width=1in,height=1.25in,clip,keepaspectratio]{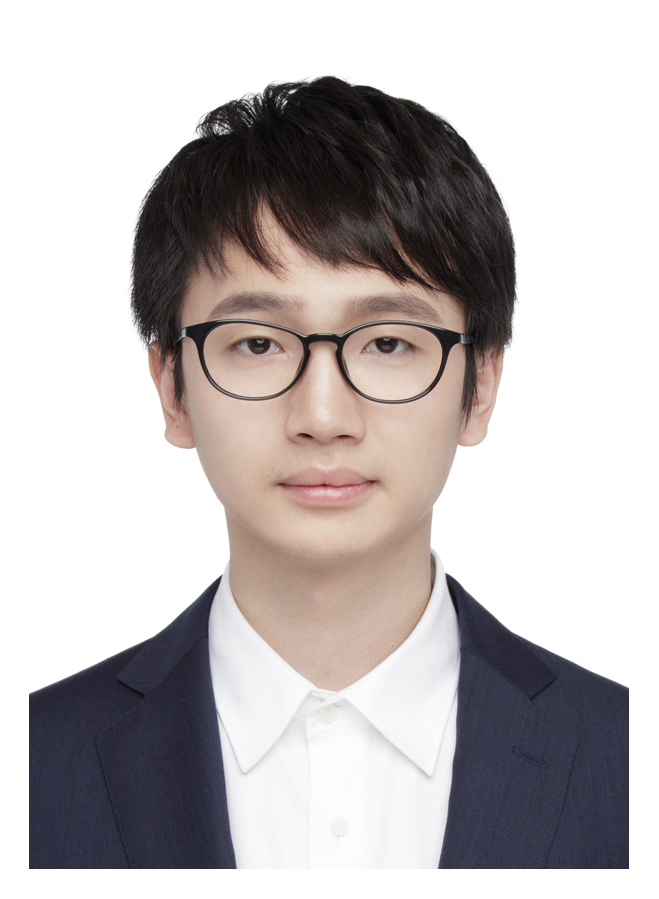}}]{Lei Fang} is currently a Research Scientist in CAAZ (Zhejiang) Information Technology Co., Ltd., Ningbo, China. His research interests mainly include GIS, remote sensing, and machine learning. \end{IEEEbiography}

\vspace{-10pt}

\begin{IEEEbiography}[{\includegraphics[width=1in,height=1.25in,clip,keepaspectratio]{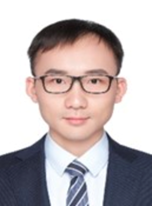}}]{Zengmao Wang} is currently  an associate professor with the School of Computer Science, Wuhan University. He has published more than 40 research papers published in the \textit{IEEE Transactions on Image Processing}, \textit{IEEE Transactions on Neural Networks and Learning Systems}, \textit{IEEE Transactions on Knowledge and Data Engineering}, CVPR, ECCV, NeurIPS, IJCAI, AAAI etc. His research interests include image processing and machine learning.\end{IEEEbiography}

\vspace{-8pt}

\begin{IEEEbiography}[{\includegraphics[width=1in,height=1.25in,clip,keepaspectratio]{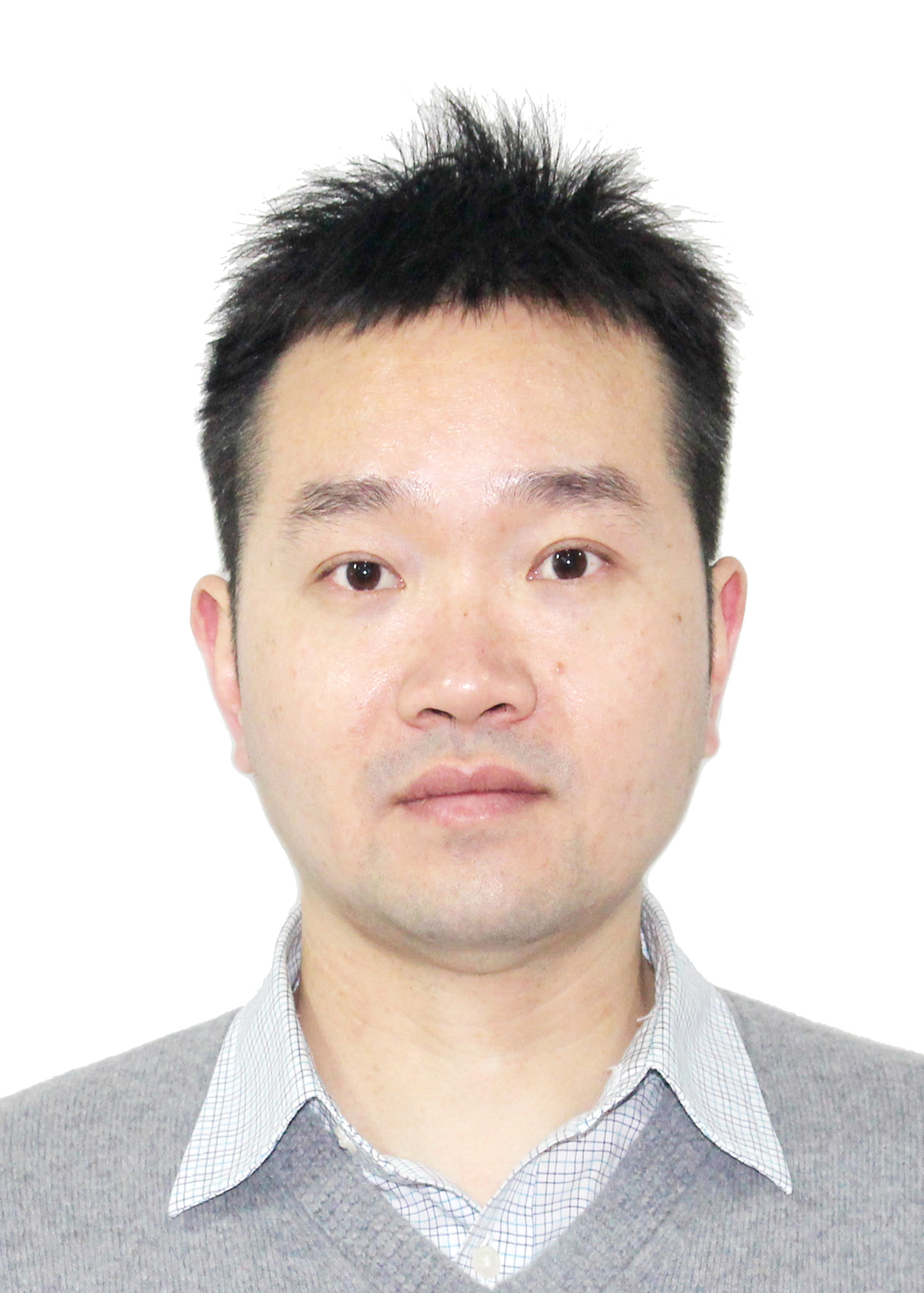}}]{Juhua Liu} is currently a professor with the School of Computer Science, Wuhan University. His research interests mainly include image processing, computer vision, natural language processing and machine learning. He has published more than 50 research papers in CV/NLP/AI, including \textit{IEEE Transactions on Pattern Analysis and Machine Intelligence}, \textit{International Journal of Computer Vision}, \textit{IEEE Transactions on Image Processing}, NeurIPS, CVPR, ACL, AAAI, IJCAI, ACM MM, \textit{etc}. \end{IEEEbiography}

\vspace{-8pt}

\begin{IEEEbiography}[{\includegraphics[width=1in,height=1.25in,clip,keepaspectratio]{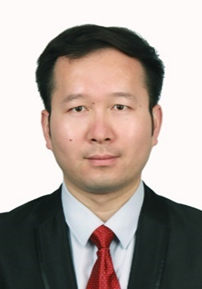}}]{Bo Du} (M'10-SM'15) is currently a professor with the School of Computer Science, Wuhan University. He has more than 100 research papers published in the \textit{IEEE Transactions on Pattern Analysis and Machine Intelligence}, \textit{IEEE Transactions on Image Processing}, \textit{IEEE Transactions on Cybernetics}, \textit{IEEE Transactions on Geoscience and Remote Sensing}, \textit{etc}. His major research interests include machine learning, computer vision, and image processing. He serves as an associate editor for \textit{Neural Networks}, \textit{Pattern Recognition}, and \textit{Neurocomputing}. He won IEEE GRSS Transactions Prize Paper Award, IJCAI Distinguished Paper Prize, IEEE Data Fusion Contest Champion, and IEEE Workshop on Hyperspectral Image and Signal Processing Best Paper Award. \end{IEEEbiography}

\vfill

\end{document}